\documentclass[letterpaper, 10 pt, conference]{ieeeconf}  

\IEEEoverridecommandlockouts                              

\usepackage{graphics} 
\usepackage{epsfig} 
\usepackage{mathptmx} 
\usepackage{times} 
\usepackage{amsmath} 
\usepackage{amssymb}  

\usepackage{comment}
\usepackage{color}
\usepackage{float}
\usepackage{cite}
\usepackage{booktabs}
\usepackage{multirow}
\usepackage{array}
\usepackage{makecell}
\usepackage{pifont}
\usepackage{algorithm}
\usepackage{algorithmic}
\usepackage{subfig}
\usepackage{amsopn}
\usepackage{bm}

\mathchardef\mhyphen="2D

\newcommand{\dplus}{Dual-SLAM$^{\bm{+}}$}
\newcommand{\Tref}[1]{Table~\ref{#1}}
\newcommand{\eref}[1]{Eq.~(\ref{#1})}

\newcommand{\Fref}[1]{Figure~\ref{#1}}

\title{\LARGE \bf
Dual-SLAM:\\ A framework for robust single camera navigation
}

\author{Huajian Huang, Wen-Yan Lin*, Siying Liu, Dong Zhang,  Sai-Kit Yeung
\thanks{Huajian Huang and Sai-Kit Yeung are with the Department of Computer Science and Engineering, Hong Kong University of Science and Technology.}%
\thanks{Wen-Yan Lin*, corresponding author, is with the School of information systems, Singapore Management University.}%
\thanks{Siying Liu is with Institute for Infocomm Research, Singapore.}%
\thanks{Dong Zhang is with the School of Electronics and Information Technology, Sun Yat-sen University.}%
}

\begin{document}

\maketitle
\thispagestyle{empty}
\pagestyle{empty}

\begin{abstract}
SLAM (Simultaneous Localization And Mapping) seeks to provide a moving agent with real-time self-localization.
To achieve real-time speed, SLAM  incrementally propagates position estimates.
This makes SLAM fast but also makes it vulnerable to local pose estimation failures.
As local pose estimation is ill-conditioned, local pose estimation failures happen regularly, making the overall SLAM system brittle. This paper attempts to correct this problem. We note that while local pose estimation is ill-conditioned, pose estimation over longer sequences is well-conditioned. Thus, local pose estimation errors eventually manifest themselves as mapping inconsistencies. When this occurs, we save the current map and activate two new SLAM threads. One processes incoming frames to create a new map and the other, recovery thread, backtracks to link  new and old maps together. 
This creates a Dual-SLAM framework that maintains real-time performance while being robust to local pose estimation failures. Evaluation on benchmark datasets shows Dual-SLAM can reduce failures by a dramatic $88\%$.    

\end{abstract}

\section{INTRODUCTION}

Simultaneous Localization And Mapping or SLAM, 
seeks to provide a moving agent with a real-time  reconstruction of
its  surroundings.
 SLAM plays a key role in   tasks such as  path planning, collision avoidance
 and self-localization.
 This paper focuses on  monocular SLAM,
 which computes the three-dimensional map from a single moving camera. While this is the most brittle of the SLAM formulations,
 it is also the most general. This means lessons learned from monocular SLAM
 can be readily applied to many other SLAM formulations. 

Monocular SLAM  works by   propagating  pose estimates  through a sequence of frames.
The propagation makes SLAM  fast. However, it also makes SLAM  brittle,
as a single  erroneous pose can create inconsistencies that destroy the map.
This problem is exacerbated  by the fact that  pose estimation on short sequences is ill-conditioned. Thus, occasional pose estimation errors are almost unavoidable.


\begin{figure}[t]
\begin{center}
\includegraphics[width=1\linewidth]{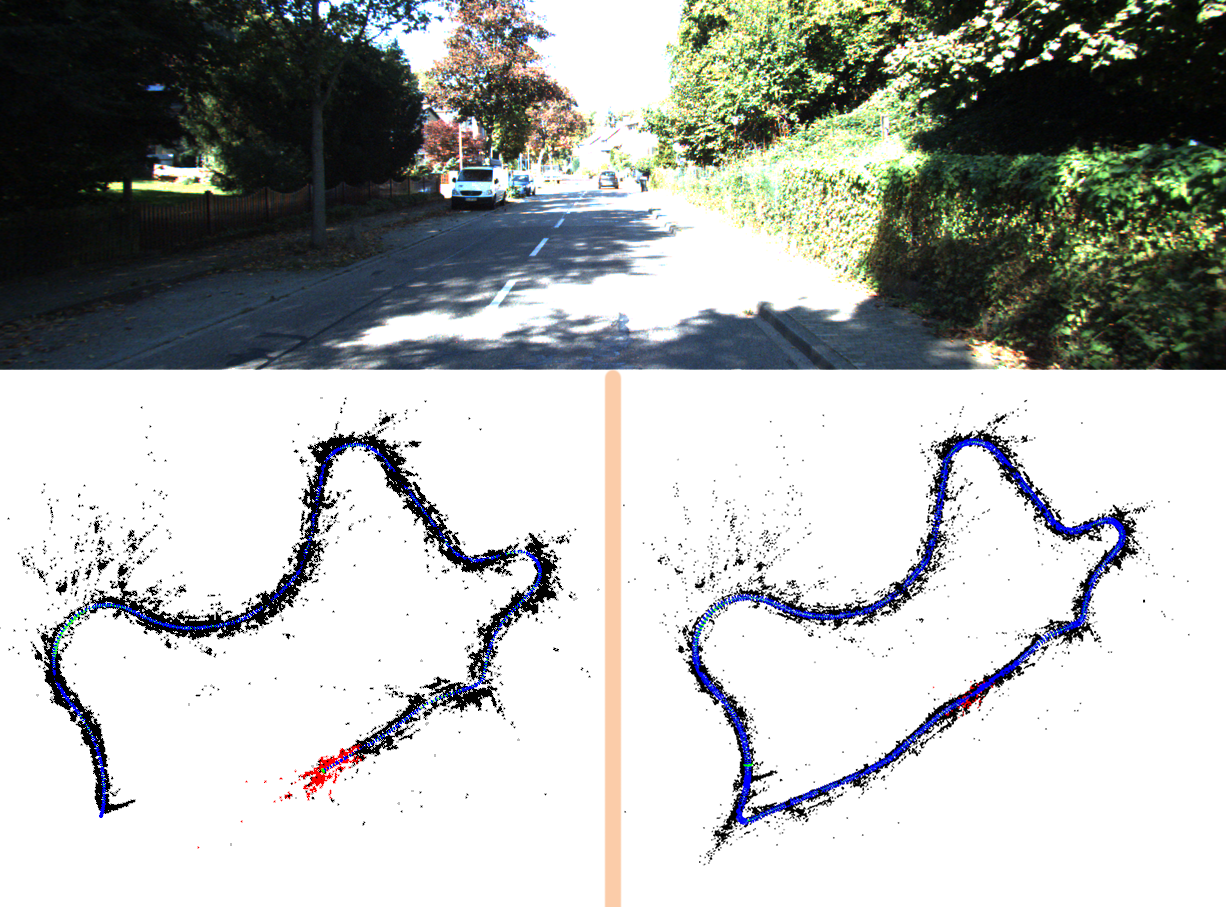}
\end{center}
   \caption{
Sequence 09 of  KITTI~\cite{KITTI}.
   Left:  ORB-SLAM's~\cite{ORB-SLAM} map.  Right:
      Dual-SLAM's map. Regions where recovery is needed are indicated in red (best viewed in color). \label{fig:intro}}
\end{figure}


Although pose estimation over short sequences (narrow baselines) is ill-conditioned, pose estimation over longer sequences
 (wide baselines) is well-conditioned.
This causes  SLAM failures to exhibit a distinctive pattern.
Extreme local errors seldom propagate indefinitely. Instead, 
the well-conditioned nature of wide baseline pose estimation means that
 if a map becomes  erroneous,
inconsistencies  with incoming  frames eventually emerge. Hence,  SLAM seldom (if ever) creates a randomly incorrect map; instead, it tends to  break when errors occur. 

Given this  failure pattern,  local pose estimation errors are easy to rectify. We demonstrate this with a  Dual-SLAM framework.
Dual-SLAM utilizes the same incremental pose estimation  as traditional SLAM. When  significant mapping inconsistencies manifest themselves, we
save the current map and initialize two new SLAMs.
One SLAM starts a new map that incorporates the incoming key-frames. The other, called recovery SLAM, propagates the new map in  the opposite direction to join with the old map,  thus by-passing the corrupted section.

This framework avoids the high computational cost  needed to prevent any possible
pose estimation failure~\cite{lin2016repmatch,lin2017code} but provides stability,
as the overall Dual-SLAM framework only  fails if 
both the  main  and recovery SLAM threads   fail simultaneously. 
This maintains  real-time performance while removing much of the frustrating brittleness plaguing monocular SLAM. Tests on standard  benchmarks 
show that Dual-SLAM can reduce failures by a dramatic $88\%$.
An example is shown in \Fref{fig:intro}.

\subsection{Related Works}

Monocular SLAM's brittleness is widely acknowledged  by the community.
To date, the primary solution is to avoid
the problem by  adding more sensing modalities.
Examples include depth sensors~\cite{klein2007parallel}, inertial measurement units~\cite{li2014lidar,achtelik2012visual,concha2016visual} and stereo cameras~\cite{ORB-SLAM2,engel2015large}. Such solutions greatly improve SLAM's robustness as
the overall  system only fails if
all modalities fail simultaneously. 
The drawback of this
 approach
  is it often results in very complicated systems. This is because
all modalities  must be calibrated  with respect to  each other.
Further,  calibration    drifts over time, thus requiring constant re-estimation~\cite{von2018direct}.
Despite its limitations, multi-modality SLAM  is
(in our opinion),  more practical than monocular SLAM.
However, the question remains. Can monocular SLAM be fundamentally stabilized?

Before attempting to  answer this question, a distinction must be made between two different SLAM problems. The first is a SLAM where
the camera is restricted to moving within a small area.
Monocular SLAM is already stable on such problems.
 Since the scene is small, the three dimensional map
can  be assumed to have been recovered accurately (there is no time to fail).
Once this is achieved,
the  SLAM  needs never fail as it can use a loop-closure detection module
to   re-localize itself with respect to the map.
 This philosophy is exploited with startling success in PTAM~\cite{klein2007parallel}.

The  second SLAM problem involves an agent roaming
a large region and rarely revisiting the same place.
Such problems are often termed video odometry.
In this case, SLAM's mapping needs to  remain  stable
over extended periods of time, without relying on  prior maps for guidance.
This is a much more difficult problem, that we seek to address with 
the  Dual-SLAM framework. 

While Dual-SLAM estimates camera pose from tracked feature points, there also exists works that employ direct pose estimation.
 Notable examples include
 LSD-SLAM~\cite{engel2014lsd} and
   LDSO~\cite{gao2018ldso} which directly use image color
   information to supplement feature tracking. 
Direct pose estimation     
    allows SLAM algorithms to make
    reasonable pose estimates  on even difficult scenes.
Their drawback is that it is hard to run  bundle-adjustment on such frameworks.
This lowers the overall quality of pose estimates.
We show that Dual-SLAM  makes traditional SLAM nearly as stable as direct pose estimation while maintaining  excellent accuracy.
More excitingly, merging Dual-SLAM with direct matching offers the possibility of extreme accuracy and stability.

Finally, we use ORB-SLAM~\cite{ORB-SLAM2} as the base SLAM for our  
Dual-SlAM framework. ORB-SLAM is itself 
 an
adaptation of  classical Structure-from-Motion  to real-time execution
on video frames.
Thus, ORB-SLAM (and by extension Dual-SlAM)  relies on many  Structure-from-Motion  modules
 like  RANSAC~\cite{fischler1981random},   feature correspondence~\cite{ORB,lowe1999object,lin2016repmatch,lin2017code},
epipolar geometry~\cite{li2006five,nister2004efficient,longuet1981computer},
  loop-closure detection~\cite{GalvezTRO12} and
 global map building ~\cite{OlsonGraph2006,geometrical}.

\begin{figure}
\begin{center}
\begin{tabular}{cc}
\includegraphics[width=0.3\linewidth]{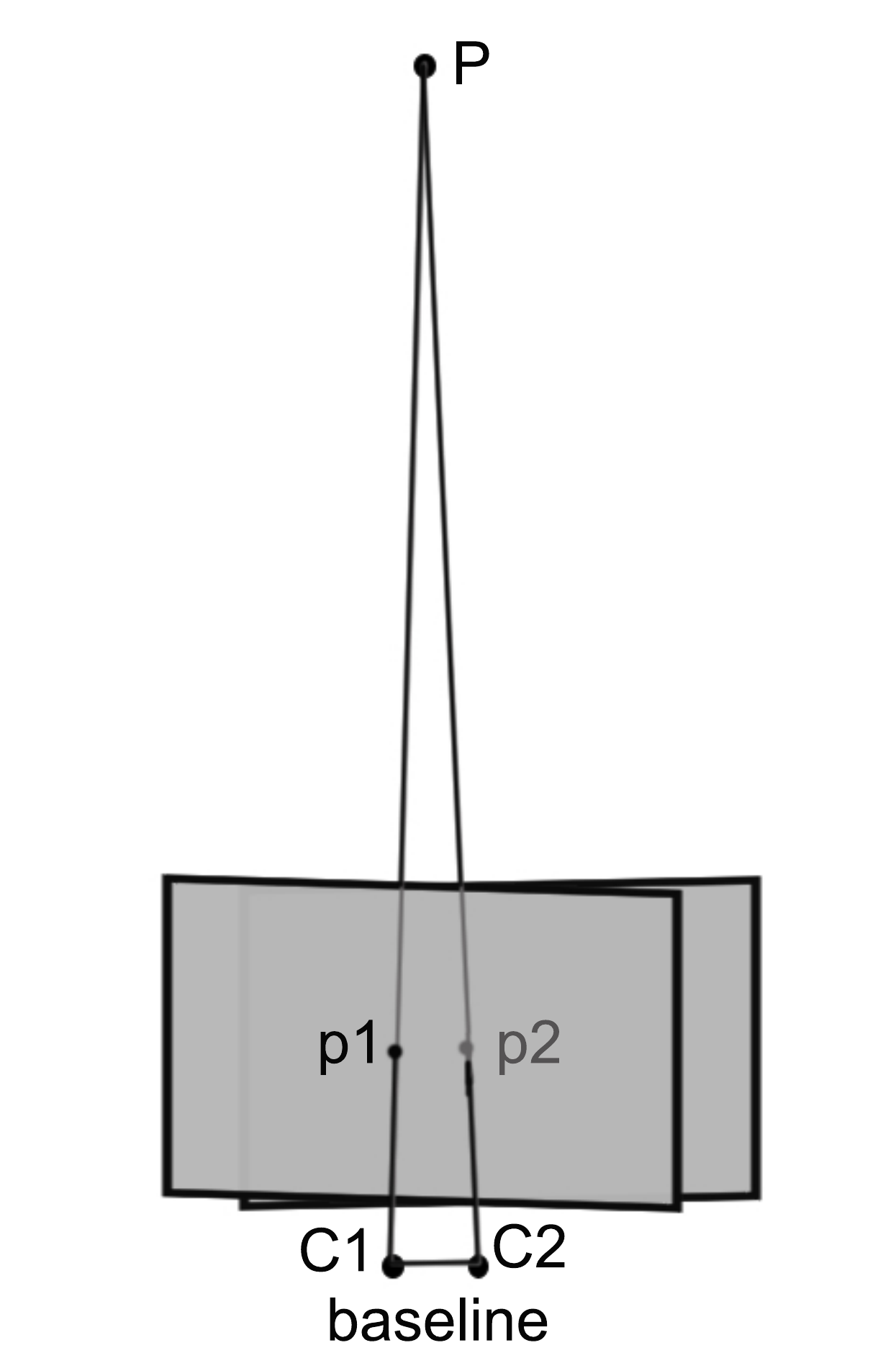}&\includegraphics[width=0.5\linewidth]{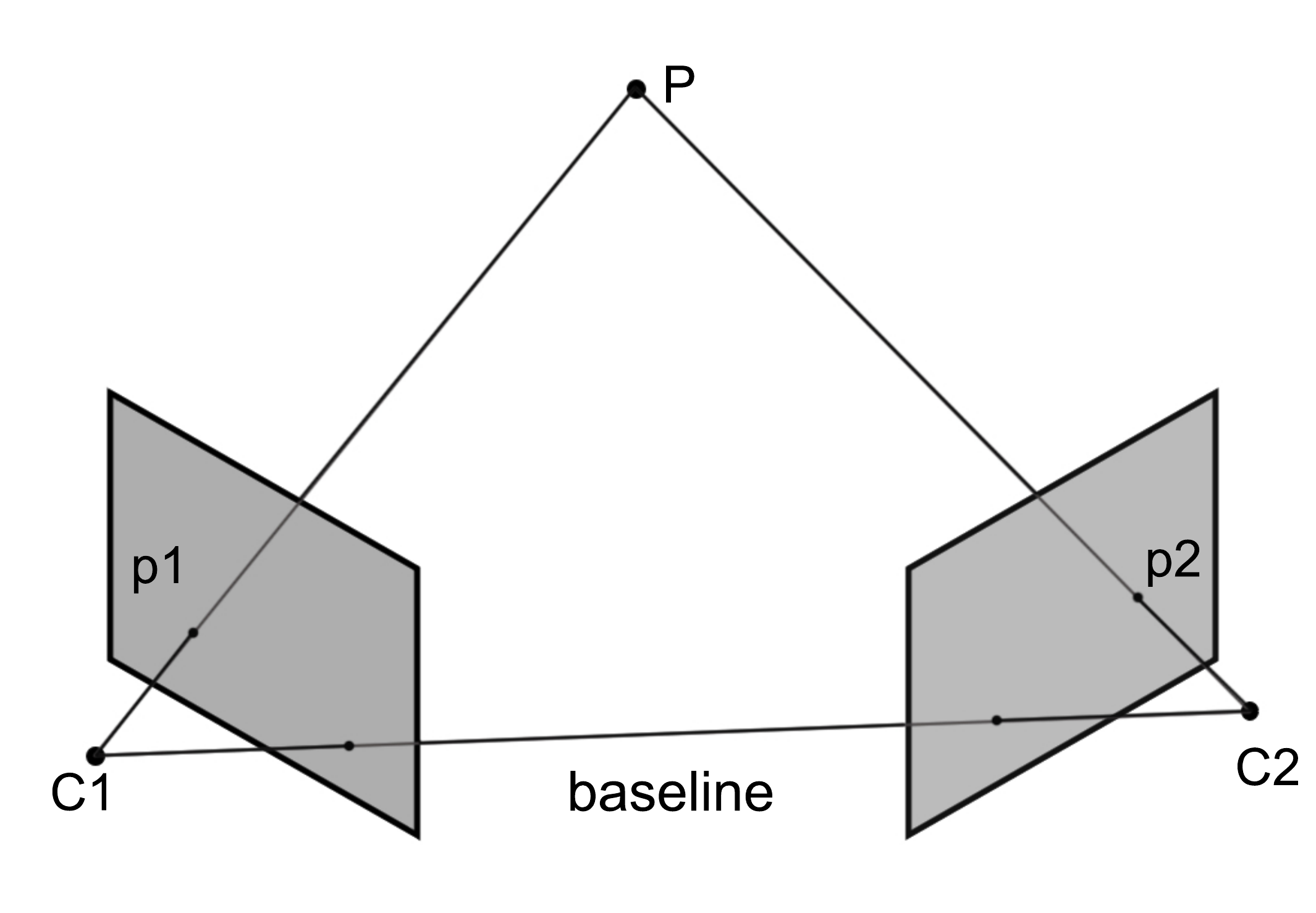}\\
narrow baseline & wide baseline
\end{tabular}
\end{center}
\caption{Pose estimation is based on triangulation. Narrow baseline triangulation
is ill-conditioned because it is based on the intersection of two near-parallel lines. In contrast, wide baseline triangulation is much better conditioned. This creates a unique SLAM failure pattern which is easy to rectify.
 }
\label{fig:illCondition}
\end{figure}

\section{OUR APPROACH}

Our approach is based on the following hypothesis regarding SLAM errors. 

To provide  real-time self-localization, SLAM incrementally propagates narrow baseline pose estimates. However, as shown in \Fref{fig:illCondition} narrow baseline pose estimation is innately ill-conditioned~\cite{knoblauch2011non}; thus,
 small errors in input (feature correspondence)  can result in large errors in the answer (pose estimates).
As such, extreme errors in local pose estimation are almost inevitable. 

These  local pose errors   create a corresponding, erroneous map that may  not be immediately  recognizable as being incorrect. However, as the baseline gets wider and triangulation better conditioned, it becomes harder for newly tracked points to be consistently fused with 
the erroneous map. This causes the SLAM sequence to break. 

If the above hypothesis is true, many breakages  that appear like tracking failures   may actually be due to stochastic errors in pose estimation. 
This motivates us to introduce the Dual-SLAM framework for enhanced SLAM robustness.

\begin{figure}[t]
\begin{center}
\includegraphics[width=0.9\linewidth]{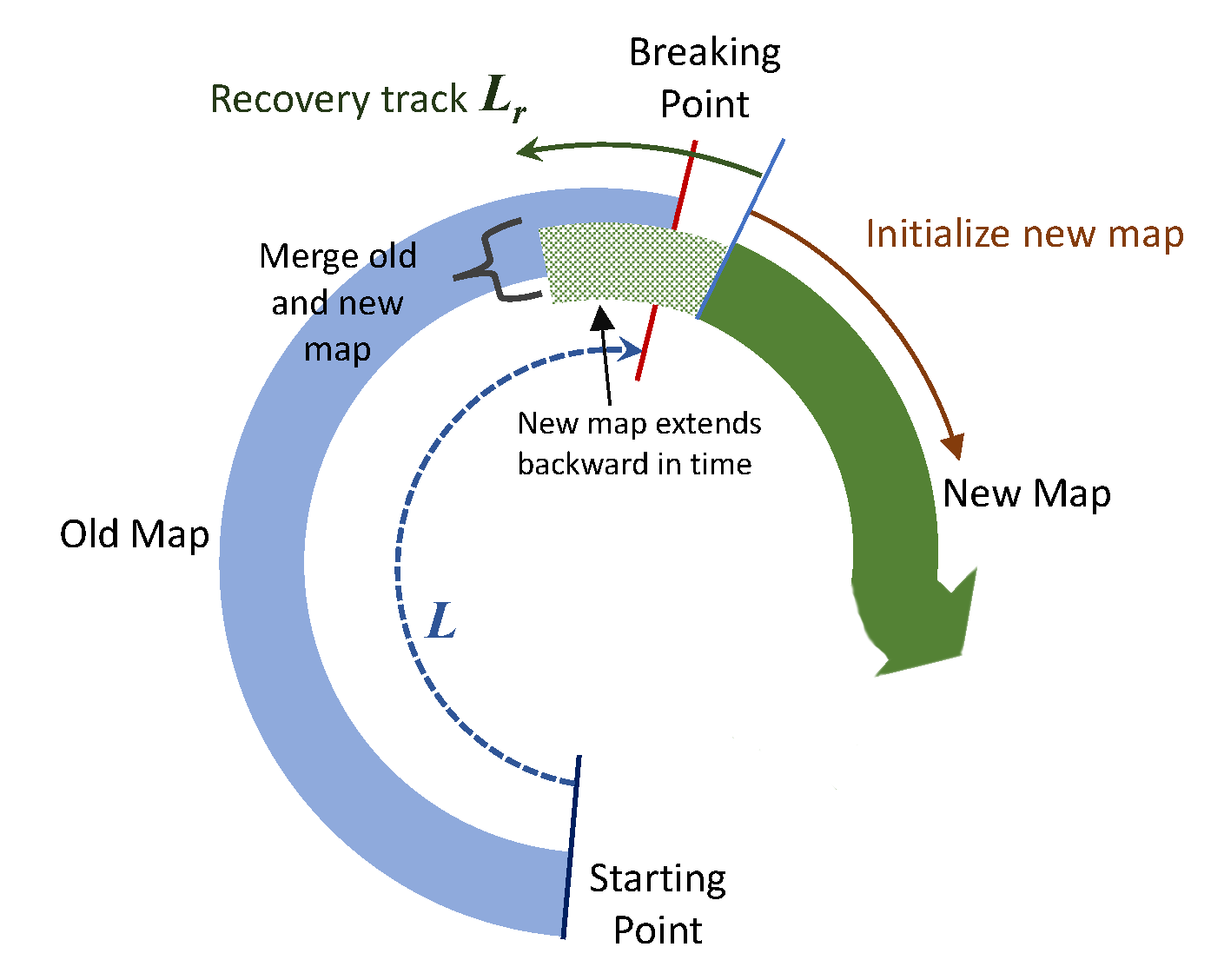}
\end{center}
\caption{Dual-SLAM handles sequence breakages by saving the existing map and  starting two SLAMs. One creates  a new map that propagates forward in time. The other SLAM propagates  map backward in time to link the new map with the old one. We show that empirically and statistically, this framework provides outstanding stability.
}
\label{fig:recovery}
\end{figure}

At its core, Dual-SLAM is  a normal SLAM with a modified handling of tracking failure. When traditional
SLAM encounters tracking failures, it typically considers the scene too difficult
and breaks. In contrast, Dual-SLAM assumes that 
tracking failures are  due to  stochastic pose estimation errors. 
Thus, instead of accepting the failure, Dual-SLAM  saves the old map and restarts a new SLAM  to create a new map from incoming frames. At the same time,  Dual-SLAM runs a recovery SLAM backward in time, to bridge the old and new maps. A schematic overview of the system is presented in \Fref{fig:recovery}. 

\subsection{Statistical Analysis of Dual-SLAM}
\label{sec:stats}

%


From a  statistical viewpoint,  Dual-SLAM  increases system reliability through redundancy.
We model a SLAM's pose estimator's failure  
 as a stochastic process. Let the probability of a SLAM system failing over a unit sequence length be $p_0$, and let the probability of  recovery thread failure be $p_r$.
As the recovery thread is run independently of the original SLAM map,  
the probability of Dual-SLAM failure with one recovery thread is, therefore,  $Pr(Failure)=p_0\times p_r$. Given $n$ independent recovery threads, the overall failure probability is given by:
\begin{equation}\label{eq:p_r}
Pr(Failure) = p_0\times(p_r)^n
\end{equation}


If both the base system and the recovery thread use the same pose estimator, we have $p_r = p_0$, since the pose estimation algorithm is considered to behave the same way in both forward and backward directions. Note that since the recovery thread can utilize information from  earlier frames stored in the memory, it is intrinsically faster. This time advantage can be exploited by using  slower but more reliable methods, such as SIFT~\cite{lowe1999object} features, in place of generic ORB features. Thus,  $p_r$ is generally lower than $p_0$.

Due to the multiplicative nature of Eq. \eqref{eq:p_r}, the failure probability can drop dramatically. Consider a case where $p_0=0.1$, $n=1$ and $p_r=0.1$. A standard SLAM would have a failure probability of $Pr(Failure)=0.1$. In contrast, Dual-SLAM with one recovery thread will have failure probability reduced to $0.01$. 
This means Dual-SLAM can survive over dramatically longer sequences than its base SLAM.
While Dual-SLAM eliminates much of the random failures  plaguing traditional SLAM, it does not solve all problems. In particular, it cannot solve intrinsically difficult cases, such as scenes with low texture. In such cases, $p_0$ and $p_r$ are close to $1$ and irrespective of the number of recovery threads, the probability of failure is still very high, i.e. $Pr(Failure) \approx 1$.

\section{SYSTEM IMPLEMENTATION}

Our base system uses {ORB-SLAM}~\cite{ORB-SLAM},
an excellent implementation of  feature-based monocular SLAM.
 {ORB-SLAM} has three primary threads. The first is a tracking thread that incrementally maps an agent's location; the second is a local-mapping thread that processes new key-frames to incrementally expand the map; the last is a loop-closure thread that reduces drift if an agent revisits a previously visited location.
{ORB-SLAM}'s tracking thread predicts camera pose using feature matches and the previous camera pose estimate. Poses are refined by a background bundle-adjustment thread and incorporated into a 3D map. When pose predictions are incorrect,
an error occurs in the 3D map; as the camera moves
onward,  incoming features are 
increasingly inconsistent with 
the now erroneous 3D map,  leading to an apparent ``tracking failure''.

Dual-SLAM primarily modifies the   handling of 
 ``tracking failures". Upon encountering a
 ``tracking failure", another tracking
thread is immediately spawned  to handle incoming frames
and initialize a new map from them. This  new
map is temporarily disconnected from the old. To reconnect them, 
 a recovery SLAM is started
 from the new map's first frame and propagates
backward in time. This extends the new map till it obtains sufficient sparse co-visibility 3D key-points with the old map for a merger.

Details of the modification are provided below.

\begin{figure}[t]
\begin{center}
\includegraphics[width=0.9\linewidth]{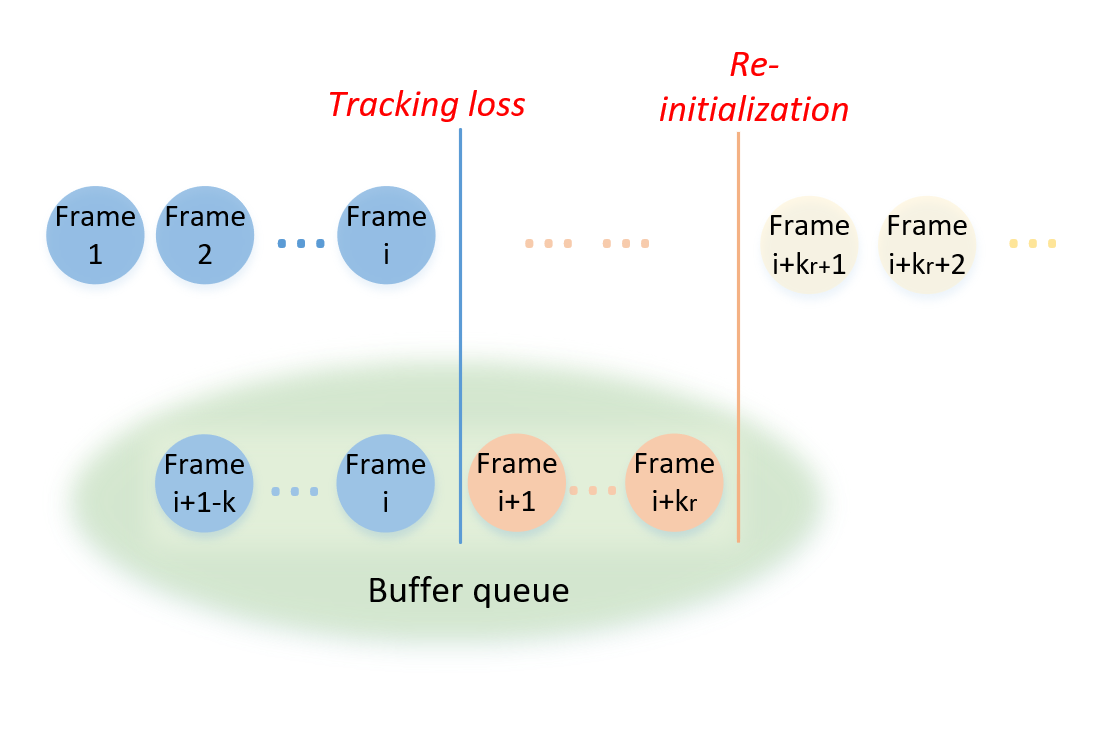}
\end{center}
\vskip -0.7cm 
\caption{The buffer queue stores frames over which the recovery SLAM propagates. The size of the window $K=k+k_r$ is adaptive.}
\label{fig_buffer}
\end{figure}

%
%

\subsection{Buffering}
\label{sect_Buffering}

To make  recovery as efficient as possible, it is necessary for
Dual-SLAM to maintain  a buffer of ORB features.  During  Dual-SLAM's normal operation, the ORB features of the most recent $k$ frames, are stored in a constantly updated buffer queue. The buffer must be  large enough to ensure an overlap with un-corrupted portions of the map; however, it cannot be so large  that storage cost becomes prohibitive. Empirically, we find  setting $k$ to the number of frames in $10$ seconds of video duration works well.

Under normal operating conditions, the buffer acts as a queue, with new frames being added and an equal number of the oldest frames being removed.  
When tracking fails, frames are no longer removed from the buffer. However, new frames are added to the buffer until the new map is re-initialized successfully. If  re-initialization occurs after $k_r$ frames, the overall buffer length is extended to  $K= k+k_r$, as shown in \Fref{fig_buffer}. Since recovery thread's SLAM is run over this buffer, the amount of time needed to recover maps is proportional to $K$.
The more delayed the  re-initialization is, the more time is needed for recovery.

\begin{algorithm}[t]
  \caption{Recovery thread \label{algo1}}

  \label{construction}
  \begin{algorithmic}[1]
  \REQUIRE  ~~\\
   buffer:  $K, F_{i+{k_r}}\rightarrow F_{i+1-k}$ ; \\
  \STATE
  \STATE $/*$ Recovery thread 1$/*$\\
  \FOR{ frame $F_j \in K $}
  \STATE pose: $p_j $ from motion model;
  \STATE tracked points:  $t_j$   conforming  to motion $p_j$;
  \IF {number of tracked points less than threshold}
  \STATE break; $/*$ Recovery thread 1 unsuccessful$/*$\\
  \ENDIF
  \ENDFOR

  \STATE Parallel running of  bundle-adjustment point cloud reconstruction using ORB-features and $p_j$ poses.
  \IF {Recovery thread 1 successful}
  \STATE \textbf{return} 3D-points
  \ENDIF

  \STATE
  \STATE $/*$ Recovery thread 2$/*$\\
  \FOR{ frame $F_i \in K$}
  \STATE $ S_j = SIFT_{match}(F_j, F_{j-1})$
  \STATE pose: $p_j = PnP(S_j)$
  \ENDFOR

  \STATE Parallel running of  bundle-adjustment point cloud reconstruction using ORB-features and $p_j$ poses.
  \IF {Recovery thread 2 successful}
  \STATE \textbf{return} 3D-points
  \ELSE 
  \STATE \textbf{return} failure
  \ENDIF

  \end{algorithmic}
  \end{algorithm}

\subsection{Recovery thread}
\label{sect_repair}

 Two different recovery threads are run sequentially. The first recovery thread is similar to a basic {ORB-SLAM}.  If the first recovery thread fails, i.e. {ORB-SLAM} is unable to propagate pose throughout the buffer,  the second recovery SLAM is 
 activated. This  SLAM uses
 SIFT~\cite{lowe1999object} features
  and a PnP~\cite{lepetit2009epnp} based pose estimation.  The procedure is summarized in algorithm~\ref{construction}.

%
%

After the recovery thread  processes all buffer frames,
 the new map will have  3D points that  overlap with the old map.
We use these common points  to fuse the  maps.

\begin{figure}[t]
\begin{center}
\includegraphics[width=0.9\linewidth]{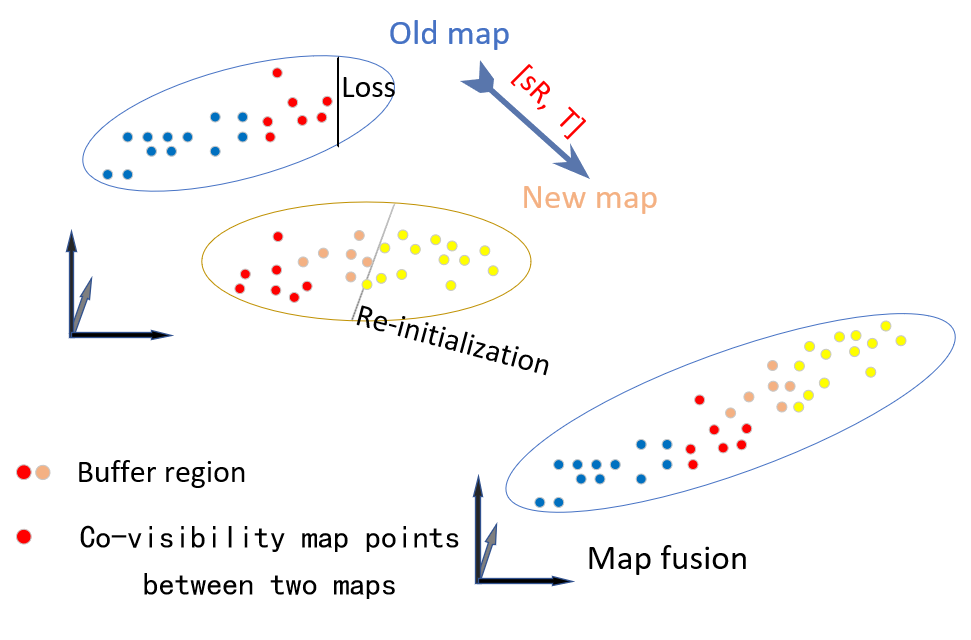}
\end{center}
\vskip -0.3cm 
\caption{Fusing old and new maps with the buffer region. }
\label{fig_mapFusion}
\end{figure}

\subsection{Map Fusion}
\label{sect_repair}

Old and new maps are related by a 3D similarity transformation,
which if estimated, will enable the fusion of the two maps. 
 As the recovery thread has extended the new map backward over the buffer, it now has 3D key-points whose locations overlap with those of the old map.
We use these 3D key-points to estimate the required similarity transformation.

Each key-point has a  corresponding key-frame ID. This means that  for  each
key-point in the new map, 
its  candidate key-point matches in the old map
can be restricted to  only those with similar   key-frame IDs. This limits the search
to a small portion of the map.

The final matching is decided on the basis of
 ORB features descriptors. The result is 
  a set of 3D-to-3D match hypotheses.
If  $\mathbf{X}_j, \mathbf{X}'_j $ are a pair of matched 3D points from the old and new maps respectively, they will be  related by
the expression:
\begin{equation}
\mathbf{X}'_j = s\mathbf{R}\mathbf{X}_j + \mathbf{T},
\end{equation}
where $\mathbf{R}$ is a $3\times 3$ rotation matrix, $\mathbf{T}$ is a $3\times1$ translation matrix,
and $s$ is a scaling factor. The similarity transformation parameters $s, \mathbf{R}, \mathbf{T}$ can be estimated by applying RANSAC with  Horn's~\cite{Horn} algorithm.
This similarity transform allows the  merger of  the maps, as shown in \Fref{fig_mapFusion}.
After the merger,  SLAM can resume normal operation.


\subsection{Enhanced version: Dual-SLAM$^{\bm{+}}$}

As mentioned  earlier in   Sec.~\ref{sect_Buffering}, rapid initialization is important to Dual-SLAM's performance. To achieve this, we modify ORB-SLAM's initialization with an advanced feature matching based on the GMS~\cite{bian2017gms} algorithm. 
We term this enhanced version \dplus.

\begin{figure}[tp]
	\begin{center}
		\includegraphics[width=1.0\linewidth]{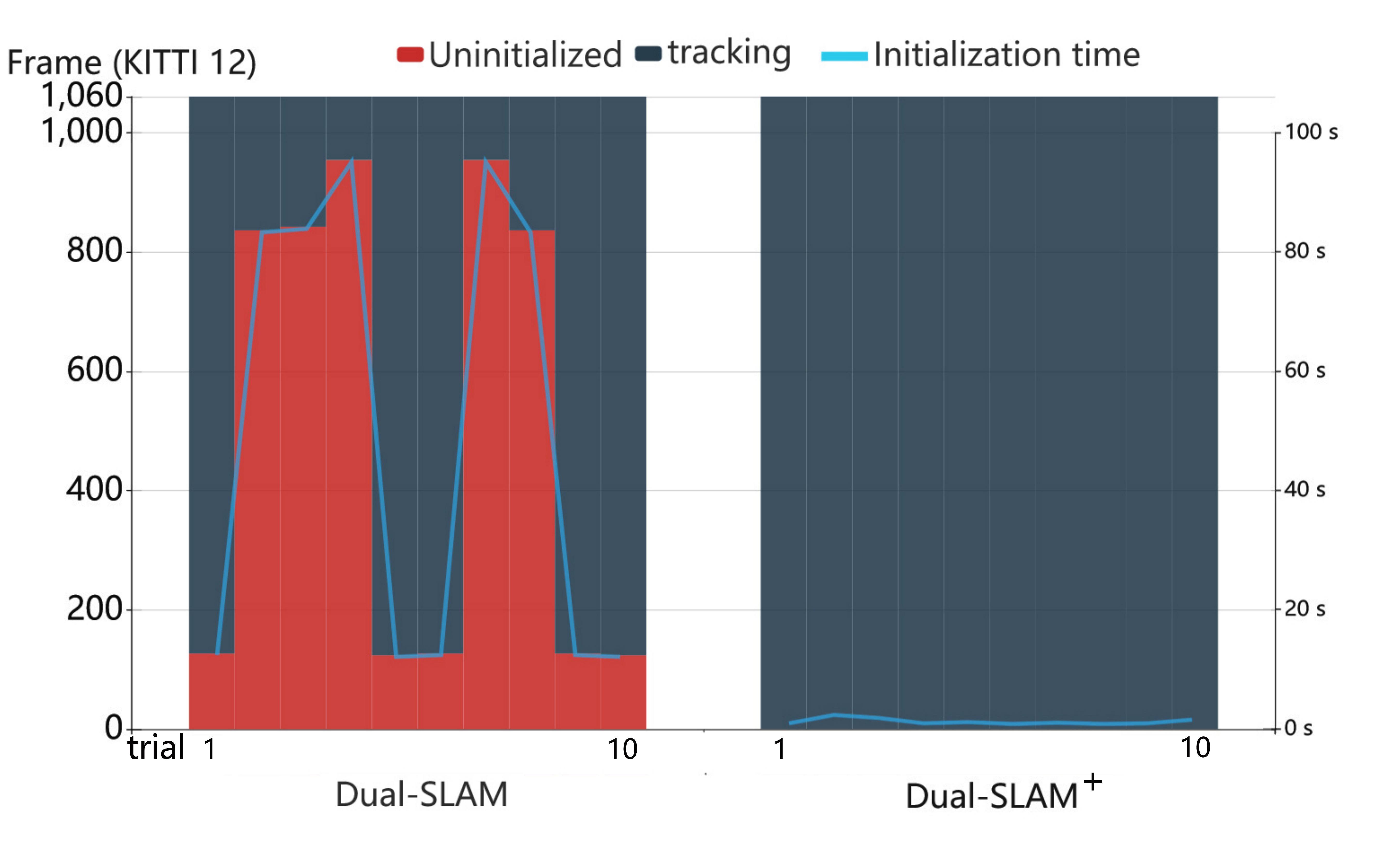}
	\end{center}
\vskip -0.3cm 
	\caption{Comparison of initialization performance on the KITTI 12. This sequence contains $1060$ frames.
		Horizontal axis:  10 different trials.
		Left vertical axis: initialization time in sequence frames numbers. Right vertical axis: initialization time (s).
		Out of 10 trials, on average,  Dual-SLAM takes $50.51$s to initialize, while Dual-SLAM$^+$ takes
		$0.5$s.}
	\label{fig_KITTI12}
\end{figure}

This modification prevents the often unacceptable lag
that occurs with basic Dual-SLAM,  
 greatly improving overall  performance 
as demonstrated in \Fref{fig_KITTI12}.

\section{EVALUATION}

We  evaluate the proposed algorithm on KITTI~\cite{KITTI} visual odometry  and TUM~\cite{engel2016monodataset}  monocular datasets.
These afford a wide variety of scenarios including outdoor, indoor and on-road scenes. Experiments are performed on an Intel Core i7-6700K laptop.
As SLAM is non-deterministic, in all experiments, we  perform $10$ trials on each sequence.

\begin{figure}[htp]
  \begin{center}
  \includegraphics[width=1\linewidth]{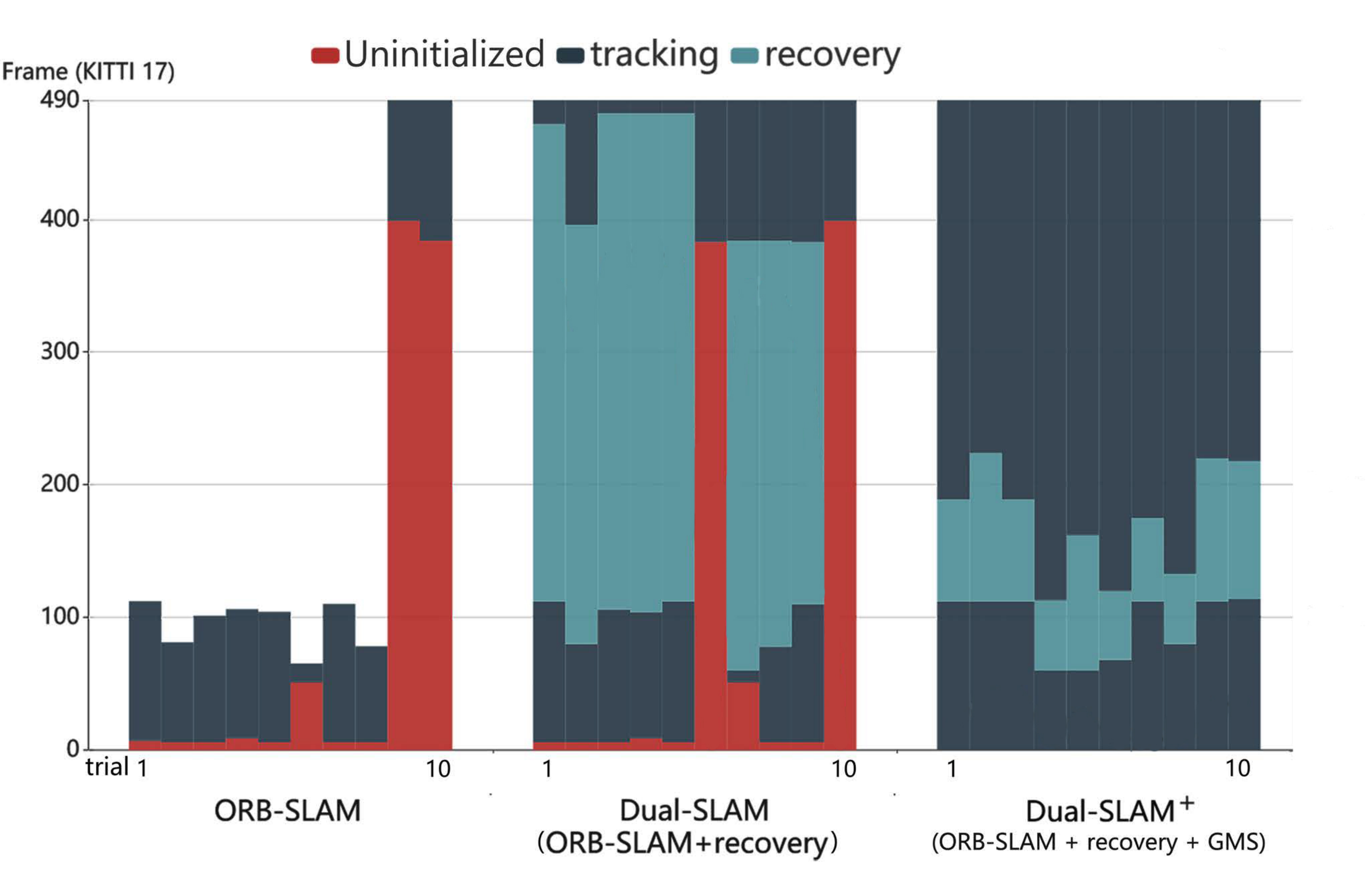}
  \end{center}
  \caption{ Comparing   SLAM algorithms on KITTI, sequence 17.
  Horizontal axis: Trial number.
  Vertical axis: Frame index. 
 ORB-SLAM  fails early in the sequence, leaving most of its profile white.   Dual-SLAM can recover 
 from the failure, providing a complete  map. However, the recovery is very slow. Finally,   \dplus demonstrates rapid recovery, making the overall navigation system more viable.}
  \label{fig_KITTI17}
  \end{figure}

\subsection{Comparisons}
We benchmark Dual-SLAM against both classic baseline SLAM systems and new state-of-the-art SLAM algorithms. {ORB-SLAM}~\cite{ORB-SLAM}, a  feature-based monocular SLAM system is used as the baseline.
This is a classic SLAM algorithm that requires no further introduction. 
   {LDSO}~\cite{gao2018ldso} represents the current state-of-the-art.
It is a sophisticated SLAM  framework that fuses traditional feature correspondence with direct image differences and 
  pose-graph optimization. We find {LDSO}'s performance is significantly better than its predecessors, {LSD-SLAM}~\cite{engel2014lsd} and {DSO}~\cite{engel2018direct}.

 In evaluations, we use ORB-SLAM at default parameters but make minor
  modifications to LDSO. 
This is because LDSO at default parameters, often terminates prematurely on easy KITTI sequences.
We find that this is due to the sensitivity of    LDSO's  loop-detection module.
Thus, to make the comparisons more meaningful,  we modify  LDSO's  loop-detection sensitivity 
for KITTI sequences.

%

\begin{table}[ tp]
  \small
  \renewcommand\arraystretch{1.2}
  \begin{center}
    \caption{  Results on the KITTI~\cite{KITTI}.  Each sequence is run 10 times
  for a total of 110 trials. The  median RMSEs of the keyframes trajectory is reported. Dual-SLAM$^+$ maintains ORB-SLAM's accuracy but is more stable.  These results are comparable with state-of-art, LDSO.\label{table_kitti}}
  \resizebox{\linewidth}{93pt}{
  \begin{tabular}{c| c c  c c  c c  c}
  \toprule
  \hline
  \multirow{2}*{Sequence} & \multicolumn{2}{c}{ \underline{ORB-SLAM} }&  \multicolumn{2}{c}{ \multirow{1}*{\underline{Dual-SLAM}} } &\multicolumn{2}{c}{ \multirow{1}*{\underline{Dual-SLAM$^+$}} }& \multirow{1}*{\underline{LDSO}}\\

     & {\footnotesize RMSE(m)}&  & {\footnotesize RMSE(m)}&  &{\footnotesize RMSE(m)}& & {\footnotesize RMSE(m)}\\
  \hline
  KITTI 00  &8.18 &$\curvearrowright$   &7.48&   & \textbf{7.30} &&9.30\\
  KITTI 01  &{$\otimes$}&{$\otimes$}  &{$\otimes$}&{$\otimes$}  & {$\otimes$} & {$\otimes$} &\textbf{10.31}\\
  KITTI 02  &24.11&  $\curvearrowright$  &24.74  & & \textbf{23.35} &&25.73\\
  KITTI 03  &\textbf{1.18}  &&1.29  & & 1.25                       &&2.14\\
  KITTI 04  & 1.65&&1.39 & & {1.20}&&\textbf{0.79}\\
  KITTI 05  & 8.23& &8.23 &  & 6.26&&\textbf{5.43}\\
  KITTI 06  & 15.84& &17.86&  & 13.29&&\textbf{12.57}\\
  KITTI 07  & {2.43}& &2.50&  & 2.59&&\textbf{1.54}\\
  KITTI 08  &51.34 &$\curvearrowright$ &50.23&  & \textbf{49.99}&&115.90\\
  KITTI 09  &50.69  &$\curvearrowright$ &13.72&  & \textbf{6.97}&&70.59\\
  KITTI 10  &7.68& &7.68&  & \textbf{6.92}&&14.79\\
  \hline
  Failure trial &\multicolumn{2}{c}{28} &\multicolumn{2}{c}{10}    &\multicolumn{2}{c}{6} &0\\
  \hline
  \multicolumn{8}{l}{$\otimes$ : the system cannot process the sequence map at all}\\
  \multicolumn{8}{l}{$\curvearrowright$ : the system has possibilities of tracking failure on this sequence}\\
  \hline
  \toprule
  \end{tabular}
  }
  \end{center}
  \end{table}

\subsection{KITTI Dataset}
The KITTI~\cite{KITTI} visual odometry dataset is captured from a vehicle.
It  includes scenes from urban and rural areas, as well as highways. 
The high speed of the data acquisition vehicle  coupled with  occasional  rapid rotations, make KITTI especially challenging. 

\begin{figure*}[h]
	\centering
	\begin{tabular}{cccc}
		\includegraphics[width=0.15\linewidth]{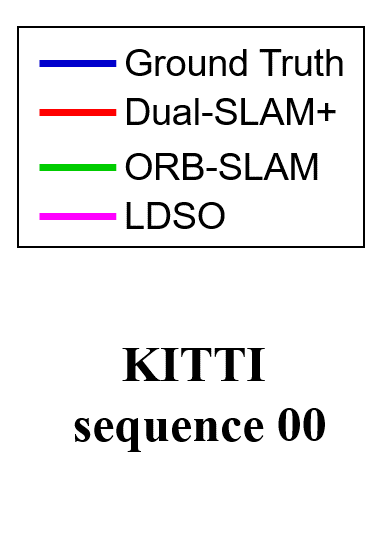} 
		&{\includegraphics[width=0.23\linewidth]{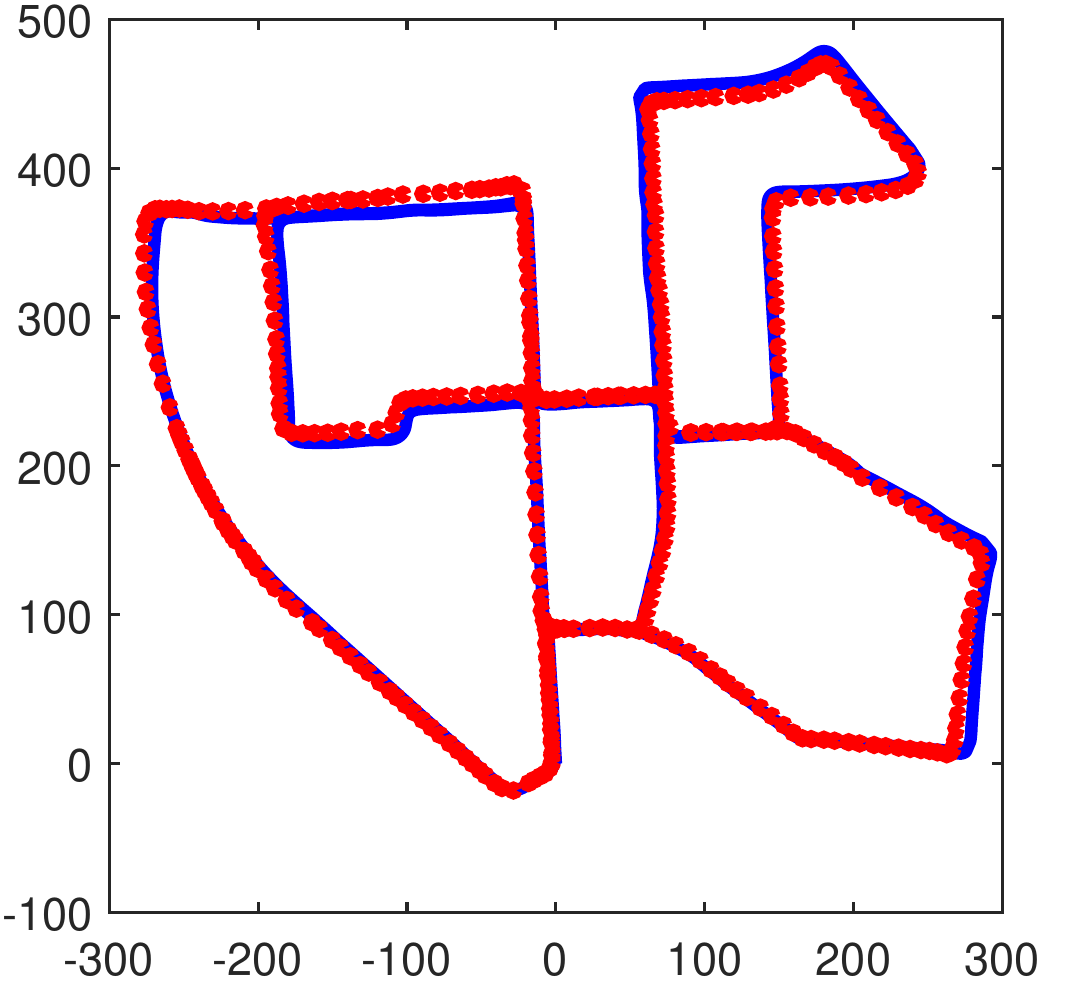} } & \includegraphics[width=0.23\linewidth]{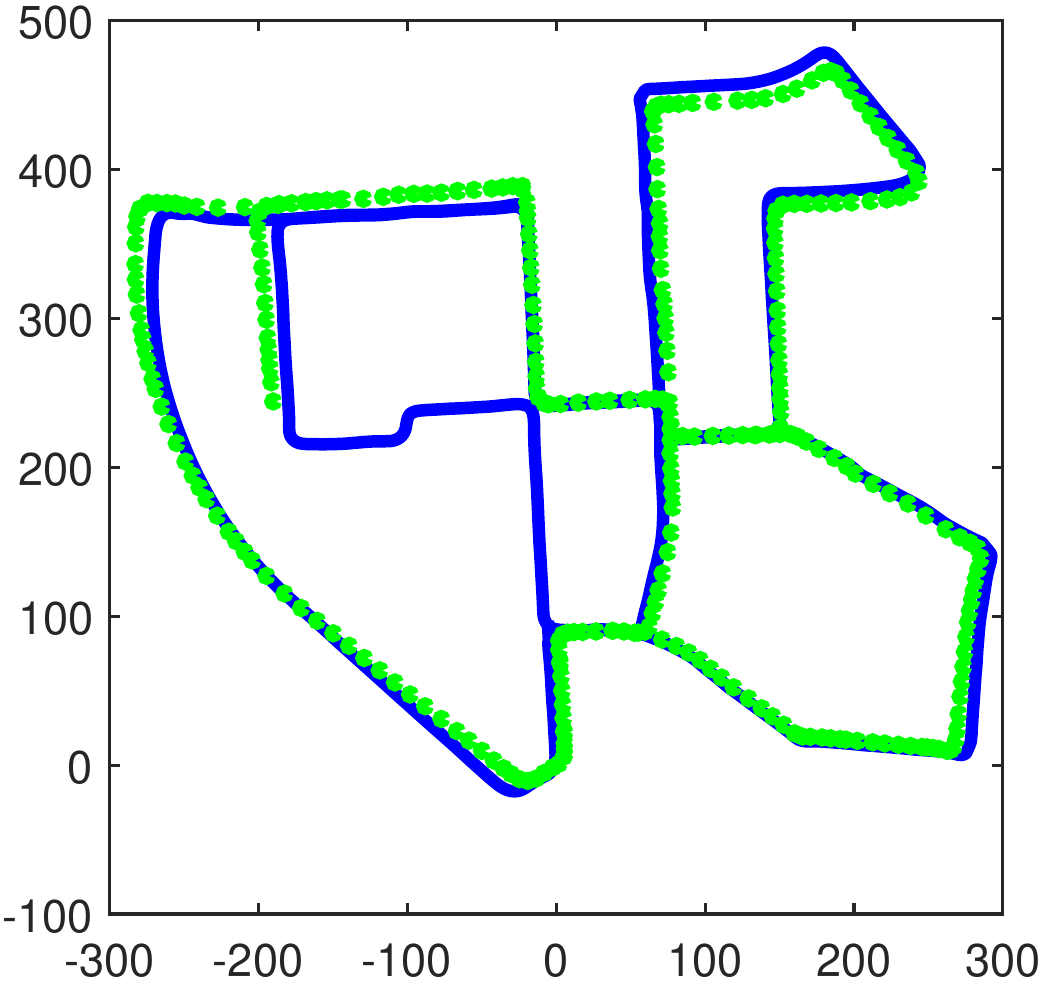} &
		\includegraphics[width=0.23\linewidth]{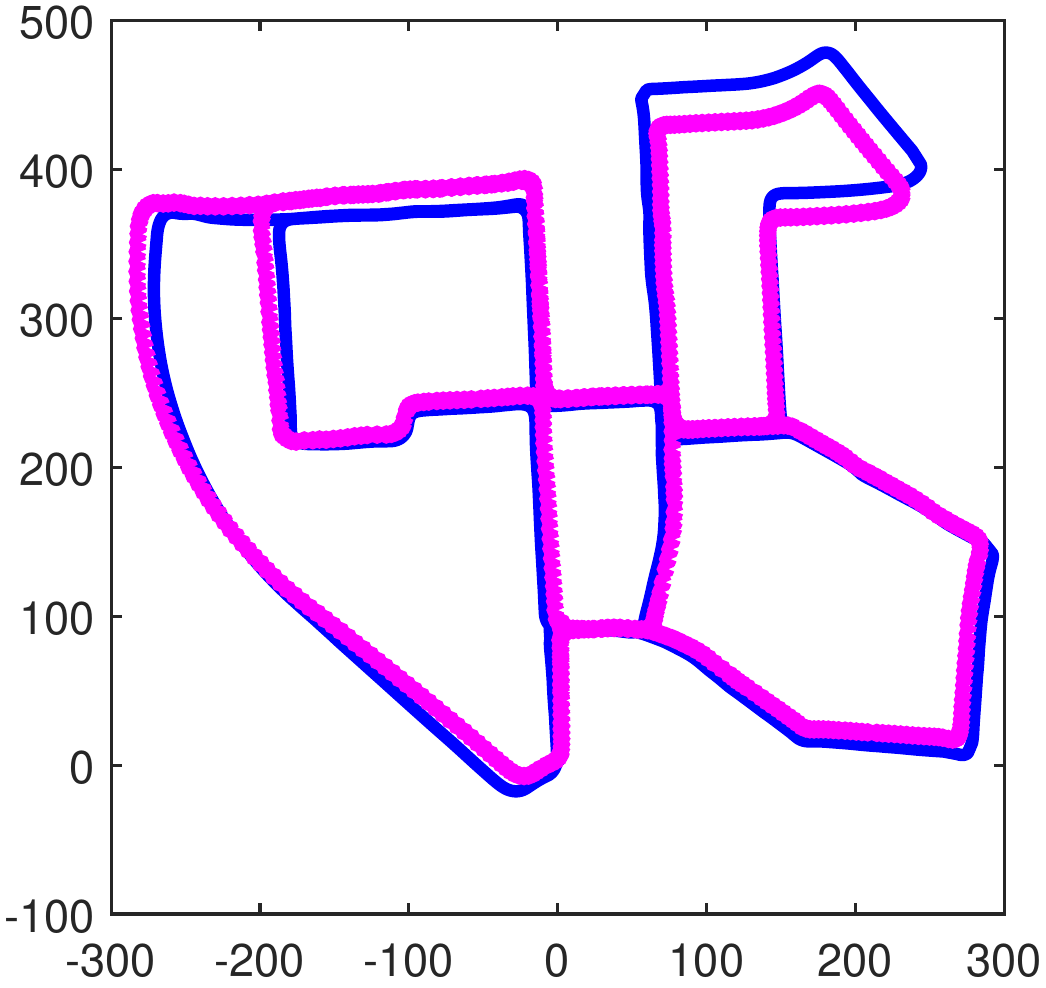} \\ \\
		\includegraphics[width=0.15\linewidth]{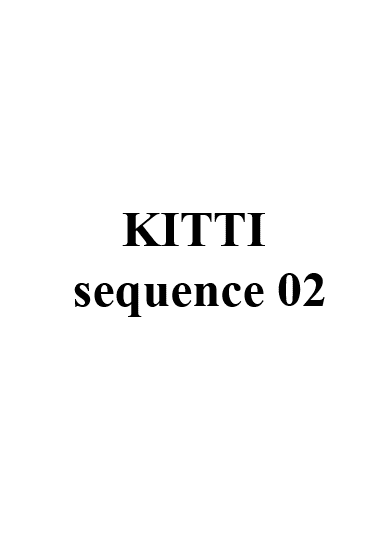} 
		& \includegraphics[width=0.18\linewidth]{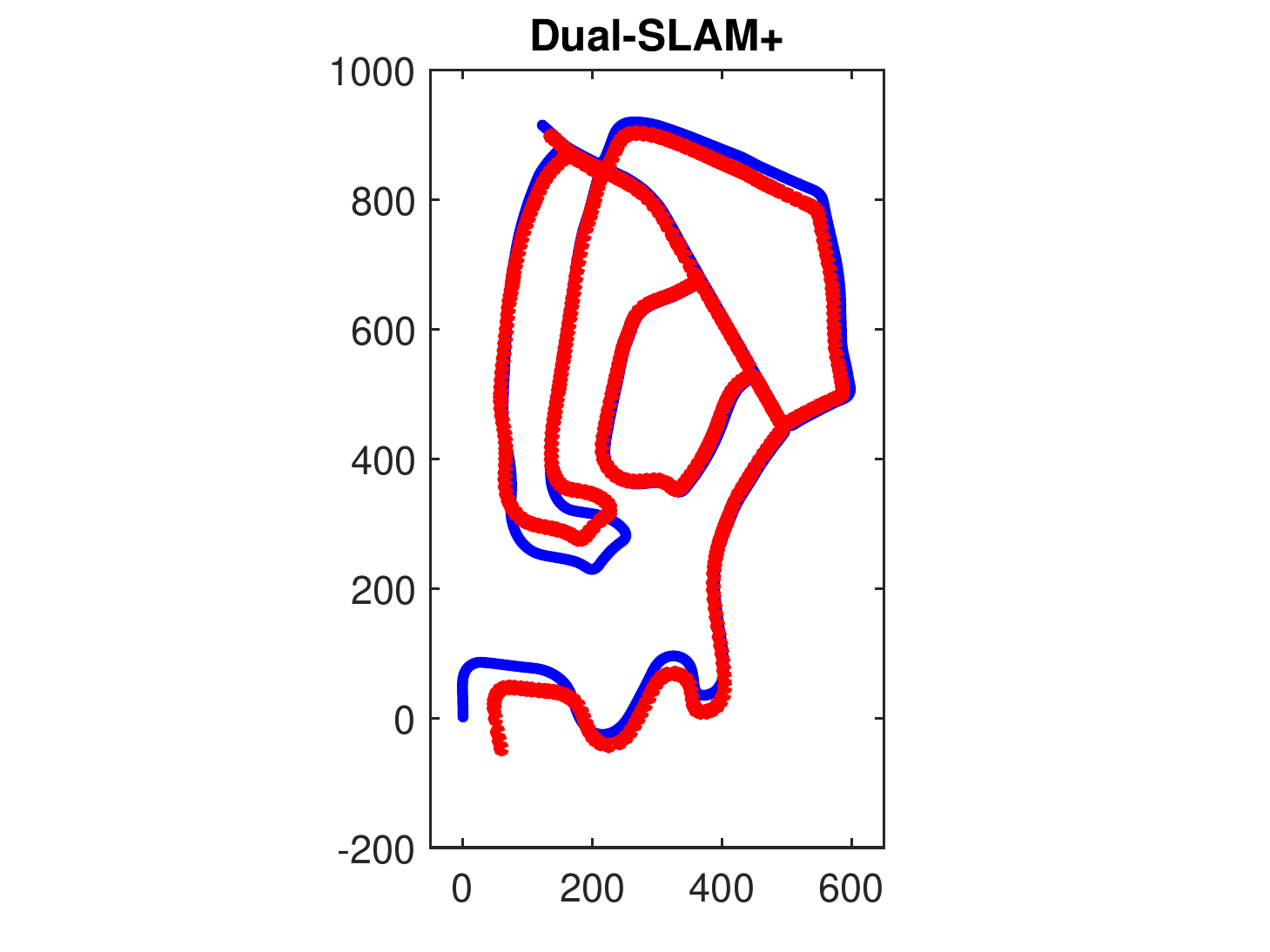} & \includegraphics[width=0.18\linewidth]{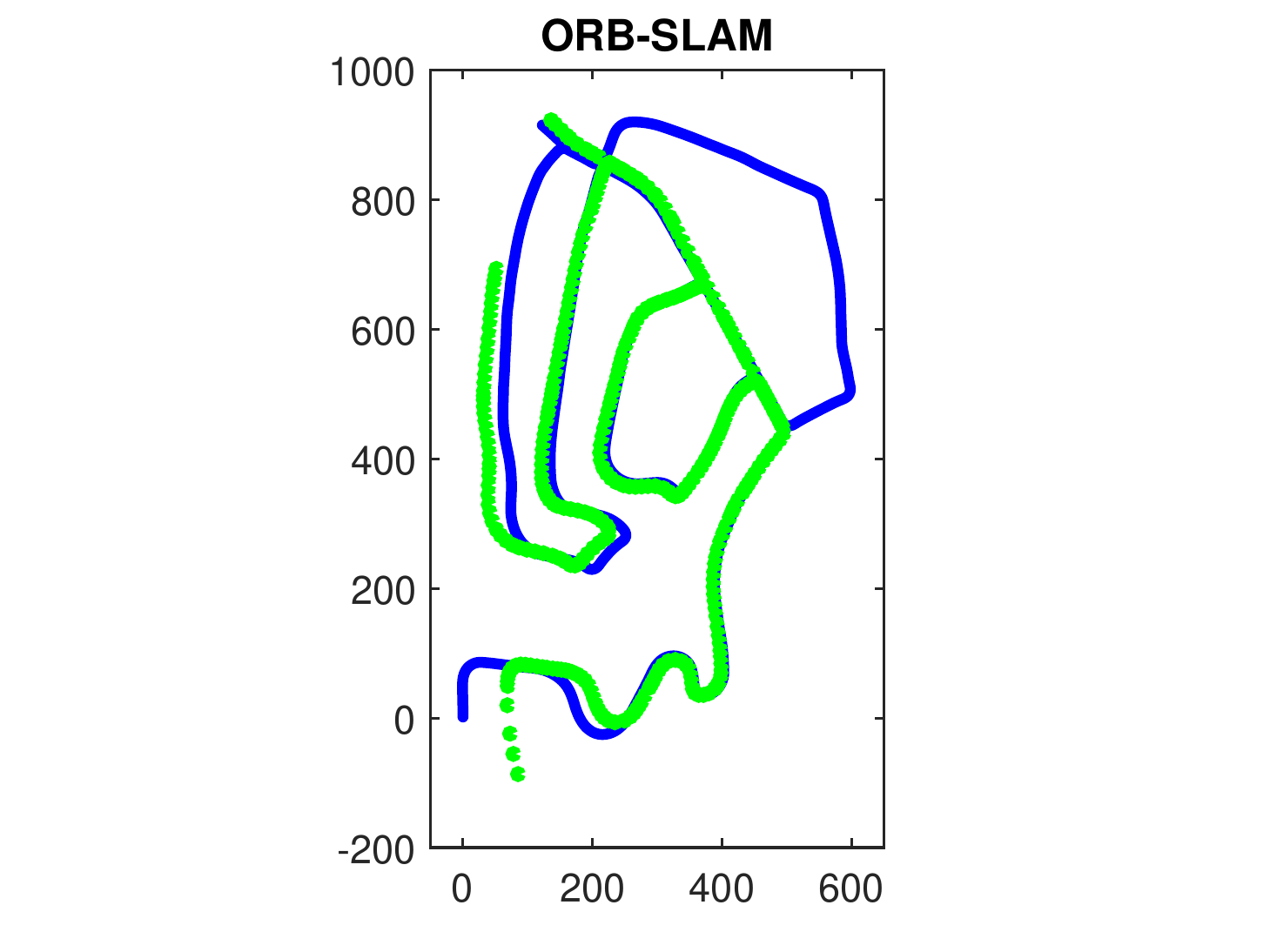} &
		\includegraphics[width=0.18\linewidth]{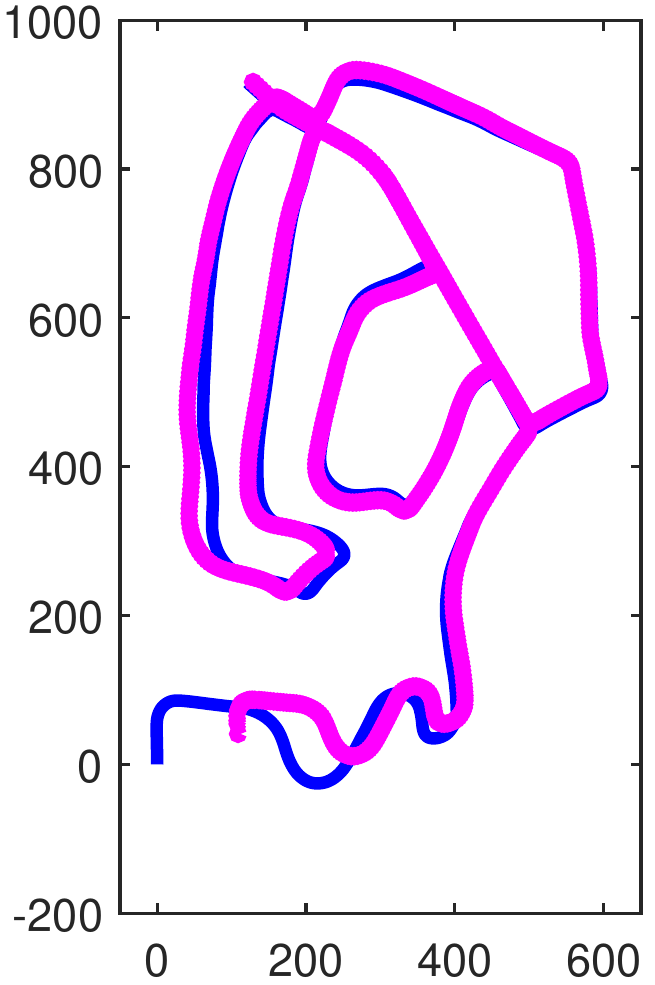} \\
		\includegraphics[width=0.15\linewidth]{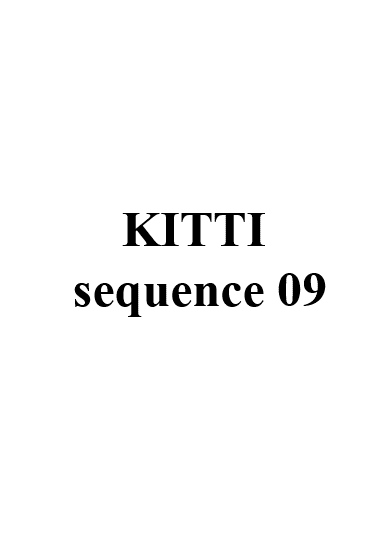}  & \includegraphics[width=0.22\linewidth]{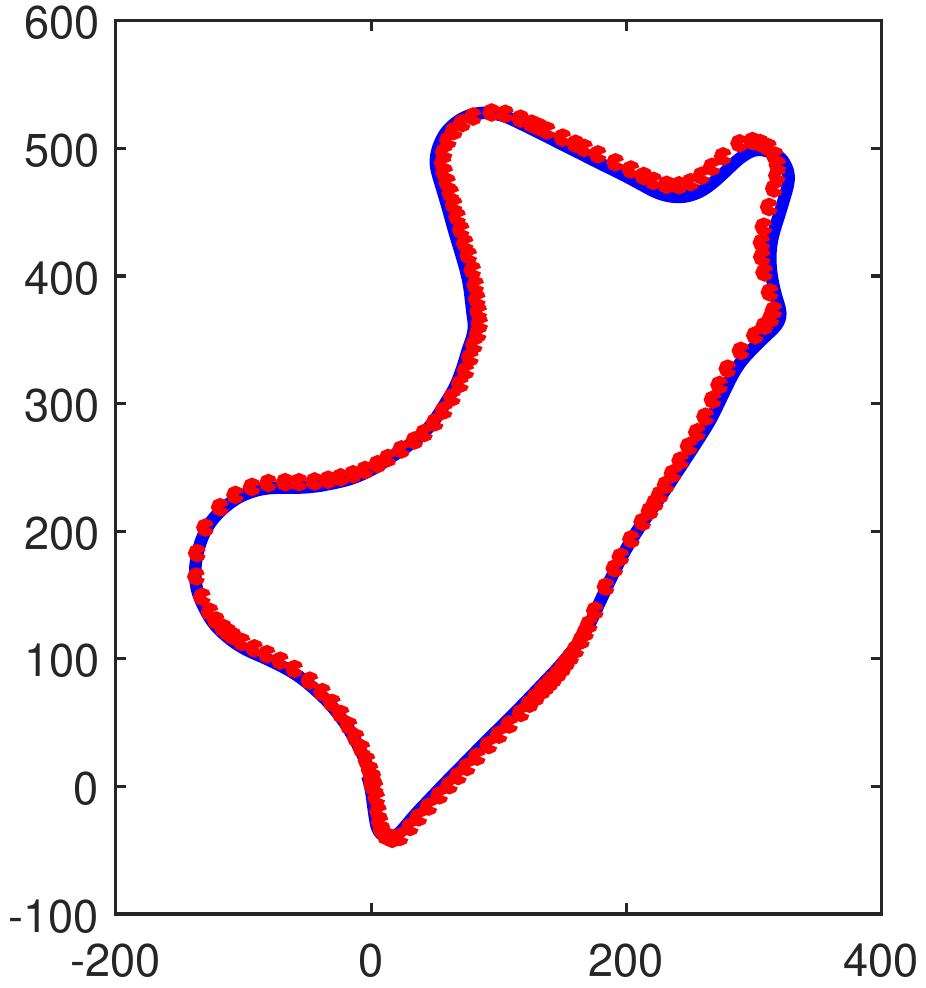} & \includegraphics[width=0.22\linewidth]{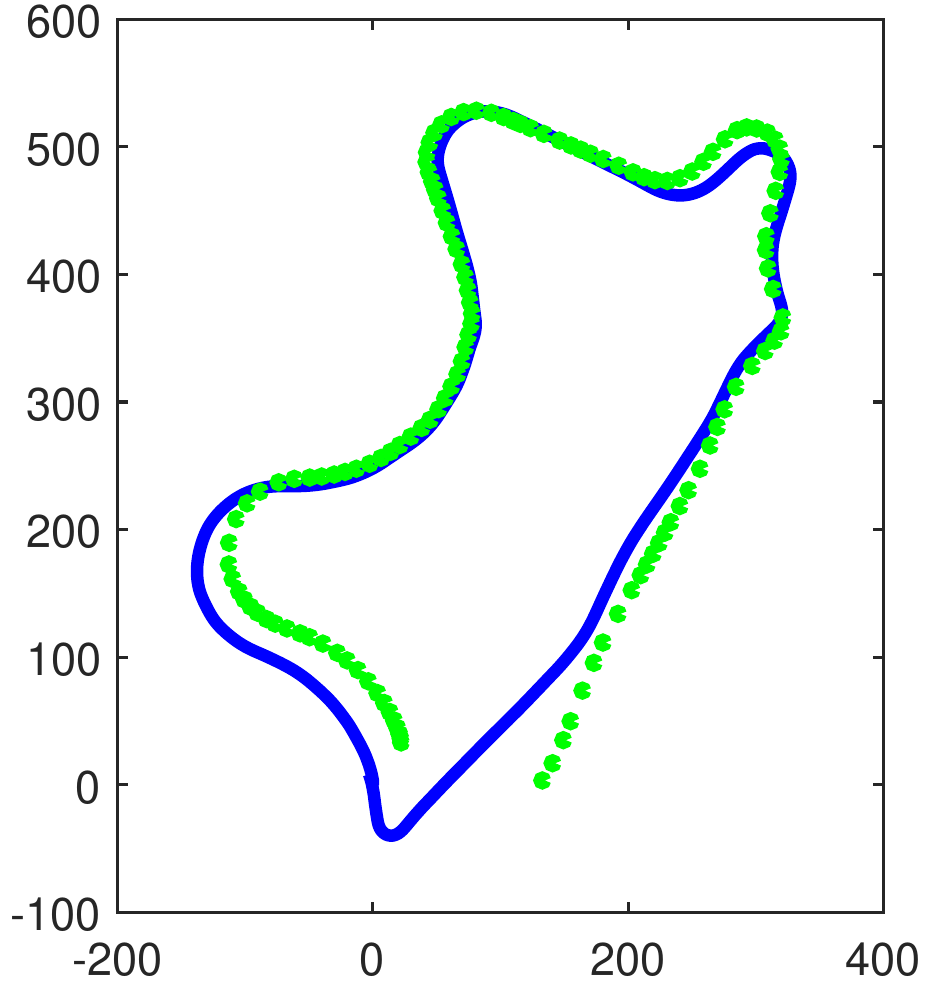} &
		\includegraphics[width=0.21\linewidth]{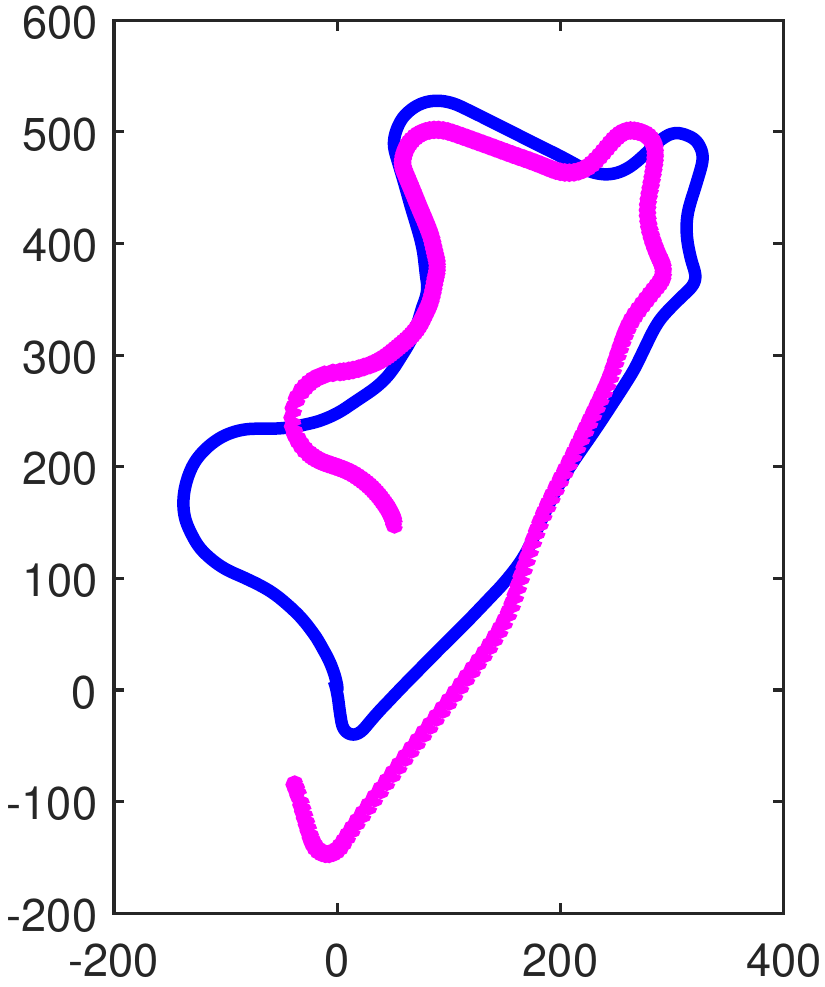} \\
		& \textbf{\dplus} & \textbf{ORB-SLAM} & \textbf{LDSO} \\
	\end{tabular}
	\caption{Overlay of SLAM and ground-truth trajectories on KITTI sequences. Ground-truth is in blue. Ideally, the SLAM trajectory would completely cover the ground-truth, making  it no longer visible. In cases
		where the SLAM breaks, large sections of ground-truth become visible. Observe that Dual-SLAM greatly improves  ORB-SLAM's results, making it competitive with LDSO. \label{fig:sy} }
\end{figure*}

The  KITTI  dataset contains 22 sequences, of which 11 sequences (00-10) possess ground truth trajectories. 
Standard ORB-SLAM  fails to construct an intact map on 5 sequences (KITTI $00$, $01$, $02$, $08$, and $09$).  Dual-SLAM reduces the failures
to just one sequence, KITTI $01$. 
The break-down of failures by individual sequences is presented in 
Tab.~\ref{table_kitti}. Observe that \dplus's improvements are so large that it makes the older (and simpler) ORB-SLAM framework competitive with state-of-the-art LDSO SLAM. 
This is most evident on sequences $08$, $09$ and $10$.  
 A  profiling of the Dual-SLAM algorithm is provided in  \Fref{fig_KITTI17}, with visual comparison on selected sequences  presented in \Fref{fig:sy}. 

\subsection{TUM-Mono Dataset}
\begin{figure}[t]
  \begin{center}
  \includegraphics[width=0.94\linewidth]{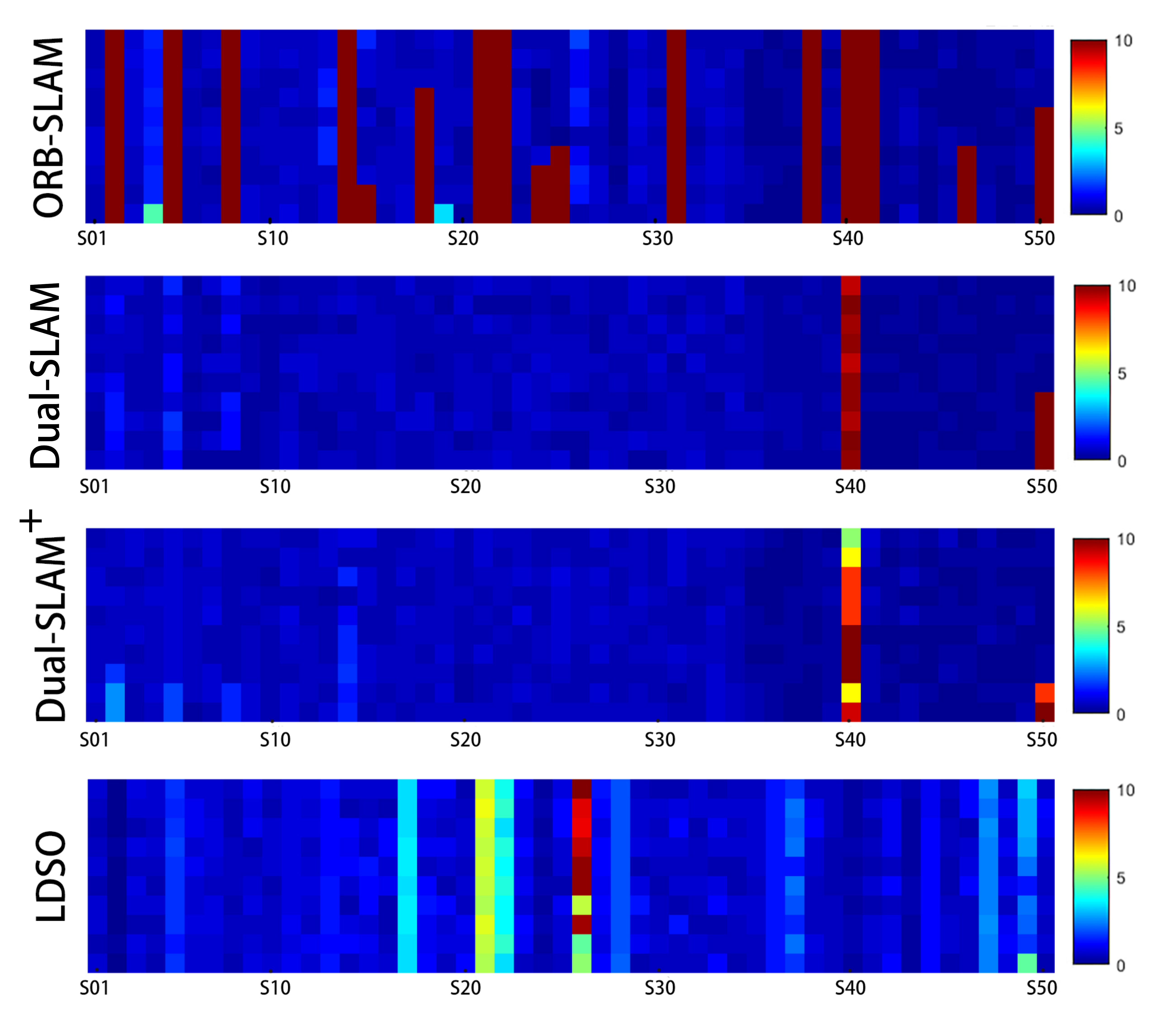}
  \end{center}
\vskip -0.3cm 
  \caption{Evaluation on TUM-Mono~\cite{engel2016monodataset}. Horizontal axis: S01-S50 represents sequence names. Vertical axis:  Results from $10$ trials. Colors represent the end-point error in meters.  Observe that traditional ORB-SLAM~\cite{ORB-SLAM} alternates between high accuracy and total failure.
  Dual-SLAM eliminates $90.47\%$ of the failures while maintaining its high accuracy.}
  \label{comparison}
  \end{figure}

TUM~\cite{engel2016monodataset} monocular visual odometry dataset contains 50 sequences, recorded in both outdoor and indoor environments. 
Ground-truth camera trajectories
are not available.   However,  as the agent moves in a loop, its start and end points  are identical. Hence,
  accuracy can be measured through end-point error. Results are tabulated in \Fref{comparison}.

\Fref{comparison} shows that 
 ORB-SLAM's performance exhibits 
 a distinctive dichotomy, with the system either
failing  completely or providing highly accurate maps.
This supports
our hypothesis that many of ORB-SLAM's errors are caused by stochastic 
 pose estimation errors, which destroy an otherwise,  very accurate map.

Dual-SLAM eliminates most of these errors. Thus, 
out of 500 trials, ORB-SLAM failed 126 times, while Dual-SLAM and Dual-SLAM$^+$ only failed 14 and 12 times respectively.
The resultant Dual-SLAM system is arguably even better than the state-of-art LDSO~\cite{gao2018ldso}. 

\begin{figure*}[t]
  \begin{center}
  \includegraphics[width=1.0\linewidth]{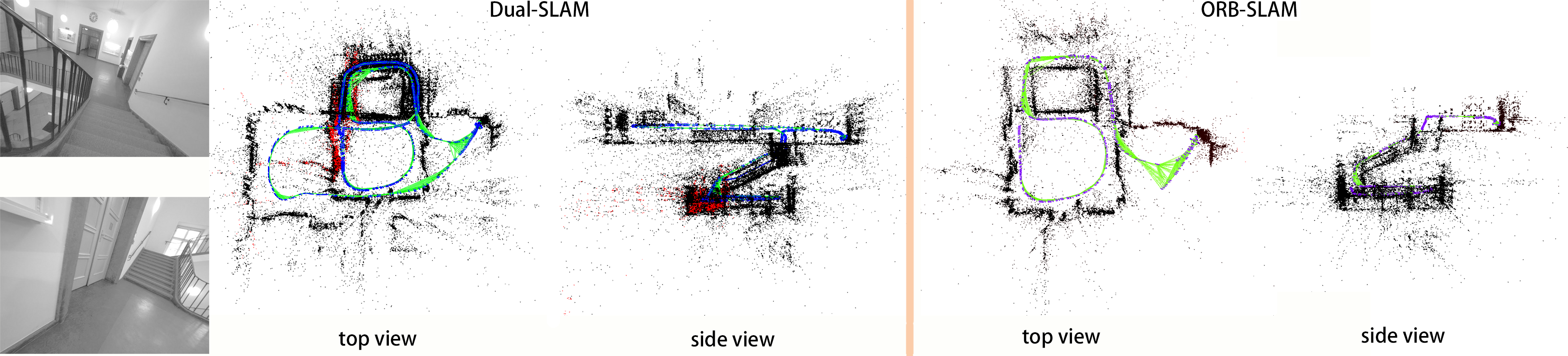}
  \end{center}
  \vskip -0.15cm 
  \caption{Sequence $41$ of TUM-Mono. Reconstructed 3D points are shown in black, key-frames in blue,
  linkages between adjacent frames
   in green and  breakage locations in red. Dual-SLAM is stable on this very difficult sequence,
   allowing it to recover much more of the map than the original ORB-SLAM.}
  \label{fig:TUMs41}
  \end{figure*}

\Fref{fig:TUMs41} shows a qualitative comparison 
on sequence 41 of TUM-Mono dataset. This is  a challenging scene, where the agent  navigates across two floors of an indoor environment.  We observe that DUAL-SLAM is 
 able to rescue breakages and perform accurate map fusion upon recovery.  
The result is a  reconstruction of a much larger floor area.

\section{ANALYSIS}
\subsection{Failure Rates}
The evaluation section suggests that Dual-SLAM is much  stabler than 
basic ORB-SLAM. \Tref {tab:Failure rate}  quantifies this,
showing   Dual-SLAM reduces  failure rates by  $84\%$-$88\%$.
This is a remarkable  result  for a relatively simple  modification.

\begin{table}[h]
	\caption{Comparing failure rates of the base ORB-SLAM with our improved Dual-SLAM. Note that  Dual-SLAM  reduces
	the number of failures sharply.}
	\begin{tabular}{c|c c c}
		&\multicolumn{3}{c}{{Failure rate = \# failure / \# trials}}\\
		\cline{2-4}
		& ORB-SLAM & Dual-SLAM & Dual-SLAM$^{\bm{+}}$\\
		\hline
		KITTI~\cite{KITTI} &28/110  &10/110 &6/110 \\
		TUM-Mono~\cite{engel2016monodataset} &126/500  &14/500  &12/500 \\
		Total & 154/610 & 24/610 & 18/610\\
		Failure Reduction & N.A. & 1-24/154 =0.84 & 1-18/154 = 0.88\\ 
	\end{tabular}
	\label{tab:Failure rate}
\end{table}

\renewcommand{\arraystretch}{1.3}
\setlength{\tabcolsep}{3pt}
 \begin{table}[ht!]
	\centering
	\caption{Ablation study of  performance on  KITTI~\cite{KITTI} and TUM-Mono~\cite{engel2016monodataset} datasets. For both datasets, we run each sequence for 10 trials and deploy a maximum of two recovery threads in case of breakage. 
We tabulate the statistics of the 	number of times recovery is needed,
the number of successful recoveries, the number of recovery threads used
in the successful cases and the average time needed for recovery. 
	}\label{tab:Recovery thread statistics}
	\begin{tabular}{c|c|c|c|c}
		\hline
		\multicolumn{2}{c|}{} & \multicolumn{2}{c|}{Dataset } & Average\\
		\cline{3-4}
		\multicolumn{2}{c|}{  }&KITTI &TUM-Mono & Recovery Time\\
		\hline
		\multirow{6}{*}{Dual-SLAM} & Total Trials  & 110     &   500    &\multirow{6}{*}{16.30s}   \\
		& Needs Recovery& 28      &   125       \\
		& Failure       & 10      &   14        \\ \cline{2-4}
		& Success       & 18      &   111       \\ 
		& With 1 thread  & 17 & 106 \\
		& With 2 threads &1  &5   \\
		\hline
		\multirow{6}{*}{\dplus}   & Total Trials  & 110     &   500    &\multirow{6}{*}{9.37s}    \\
		& Needs Recovery& 30      &   126       \\
		& Failure       & 6      &   12        \\ \cline{2-4}
		& Success       & 24      &   114       \\
		& With 1 thread & 22      &   110 \\
		& With 2 threads & 2      &    4   \\
		\hline
	\end{tabular}

\end{table}


 \subsection{Ablation Study}


\Tref{tab:Recovery thread statistics} provides a break down of performance statistics  based on the initialization (Dual-SLAM vs \dplus) and the number of recovery threads deployed. 
  Dual-SLAM and \dplus \hskip 0.1cm deployed  recovery threads $153$ and $156$ times respectively. Successful recovery occur in $84.3\%$ and $88.5\%$ of the cases. 
Of these, in $95.3\%$ and $95.7\%$ of the cases, recovery was successful at the  first  attempt. 


The high success of the first  recovery thread 
has  practical and theoretical implications. 
At a practical level, this phenomenon
makes  recovery much faster, enhancing  Dual-SLAM's viability as
a  navigational technique. At a theoretical level, it  validates  \eref{eq:p_r}
which predicts that a small amount of redundancy is sufficient to  significantly reduce the likelihood of failure.

\subsection{Relation to RANSAC}

RANSAC is another algorithm that relies on stochastic  pose prediction.  In RANSAC, 
a small number of putative correspondences are used to estimate the pose between two frames. This process
is iterated $N$ times and 
the pose that is in agreement with the largest number of correspondences, is 
 assumed  to be the correct pose estimate.

Despite the superficial similarity between
 RANSAC and Dual-SLAM, more RANSAC iterations  cannot solve the ill-conditioning of narrow baseline
pose estimation as  ill-conditioning implies the global minimum of a RANSAC cost may correspond to a highly incorrect pose estimate. 
Dual-SLAM addresses this problem by  considering pose solutions over   much wider baselines.

\subsection{Framework Limitations}
Dual-SLAM has two  primary
weaknesses. 
The first  is its assumption that mapping errors always  lead to  breakages. 
As explained earlier, the well-conditioned nature
of 
 wide-baseline pose estimation makes this a good assumption. 
However, it is occasionally violated. When this occurs, the   
resultant maps often exhibit 
large drift errors.
Note that in these cases the problems are not caused by the introduction of   new errors
but  the result of  a failure to correct existing errors. Examples of such failures are present in
sequence 01 of  KITTI and 
sequence 50 of  TUM-Mono.

Dual-SLAM's other major   weakness is its vulnerability to low textured scenes.  This
causes problems for all feature-based SLAM systems and  Dual-SLAM is no exception.
An example of this case occurs in sequence 40 of TUM-Mono.

\subsection{Relation to LDSO and Future Work}

Results in \Tref{table_kitti} and \Fref{comparison} show that overall, Dual-SLAM's
performance is comparable to  LDSO. However, closer inspection reveals that the two techniques 
are very different, with each one  favoring a different scene type. For general scenes,   Dual-SLAM is typically more  stable and accurate. However, LDSO  
has an advantage on   scenes with extremely low textures or high motion blur. 
   
These strengths and weaknesses are the result of  different design philosophies. Dual-SLAM
assumes 
 tracking errors are the result of pose estimation failures, which
it attempts to correct through re-estimation. In contrast, LDSO assumes
tracking failures are the result of insufficient features, which it attempts to rectify 
trough dense matching. 

We believe that neither the Dual-SLAM or LDSO approach is fully correct and 
that the ideal SLAM would be one that fuses both of these design philosophies. 
This offers the exciting potential for further performance improvements.

\section{CONCLUSION}
This paper hypothesizes that the source  of monocular
SLAM brittleness is  stochastic errors incurred by  narrow baseline pose estimations.  We develop a Dual-SLAM framework to address this problem, creating  a simple but general  framework for enhanced SLAM performance.

\section*{Acknowledgment}\vskip -0.05cm
This research is supported by the Singapore Ministry of Education (MOE) Academic Research Fund (AcRF) Tier 1 grant
 and internal grant from HKUST(R9429). We also  thank Weibin Li and Miaoxin Huang
  for their generous  help. 

\bibliographystyle{IEEEtran}
\bibliography{IEEEabrv,egbib}

\begin{thebibliography}{10}
\providecommand{\url}[1]{#1}
\csname url@rmstyle\endcsname
\providecommand{\newblock}{\relax}
\providecommand{\bibinfo}[2]{#2}
\providecommand\BIBentrySTDinterwordspacing{\spaceskip=0pt\relax}
\providecommand\BIBentryALTinterwordstretchfactor{4}
\providecommand\BIBentryALTinterwordspacing{\spaceskip=\fontdimen2\font plus
\BIBentryALTinterwordstretchfactor\fontdimen3\font minus
  \fontdimen4\font\relax}
\providecommand\BIBforeignlanguage[2]{{%
\expandafter\ifx\csname l@#1\endcsname\relax
\typeout{** WARNING: IEEEtran.bst: No hyphenation pattern has been}%
\typeout{** loaded for the language `#1'. Using the pattern for}%
\typeout{** the default language instead.}%
\else
\language=\csname l@#1\endcsname
\fi
#2}}

\bibitem{KITTI}
A.~Geiger, P.~Lenz, and R.~Urtasun, ``Are we ready for autonomous driving? the
  kitti vision benchmark suite,'' \emph{IEEE Conference on Computer Vision and
  Pattern Recognition (CVPR)}, vol. 157, no.~10, pp. 3354--3361, 2012.

\bibitem{ORB-SLAM}
R.~Mur-Artal, J.~M.~M. Montiel, and J.~D. Tardos, ``Orb-slam: a versatile and
  accurate monocular slam system,'' \emph{IEEE Transactions on Robotics},
  vol.~31, no.~5, pp. 1147--1163, 2015.

\bibitem{lin2016repmatch}
W.-Y. Lin, S.~Liu, N.~Jiang, M.~N. Do, P.~Tan, and J.~Lu, ``Repmatch: Robust
  feature matching and pose for reconstructing modern cities,'' in
  \emph{European Conference on Computer Vision}.\hskip 1em plus 0.5em minus
  0.4em\relax Springer, 2016, pp. 562--579.

\bibitem{lin2017code}
W.-Y. Lin, F.~Wang, M.-M. Cheng, S.-K. Yeung, P.~H. Torr, M.~N. Do, and J.~Lu,
  ``Code: Coherence based decision boundaries for feature correspondence,''
  \emph{IEEE transactions on pattern analysis and machine intelligence},
  vol.~40, no.~1, pp. 34--47, 2017.

\bibitem{klein2007parallel}
G.~Klein and D.~Murray, ``Parallel tracking and mapping for small ar
  workspaces,'' in \emph{Proceedings of the 2007 6th IEEE and ACM International
  Symposium on Mixed and Augmented Reality}.\hskip 1em plus 0.5em minus
  0.4em\relax IEEE Computer Society, 2007, pp. 1--10.

\bibitem{li2014lidar}
R.~Li, J.~Liu, L.~Zhang, and Y.~Hang, ``Lidar/mems imu integrated navigation
  (slam) method for a small uav in indoor environments,'' in \emph{2014 DGON
  Inertial Sensors and Systems (ISS)}.\hskip 1em plus 0.5em minus 0.4em\relax
  IEEE, 2014, pp. 1--15.

\bibitem{achtelik2012visual}
M.~W. Achtelik, S.~Lynen, S.~Weiss, L.~Kneip, M.~Chli, and R.~Siegwart,
  ``Visual-inertial slam for a small helicopter in large outdoor
  environments,'' in \emph{2012 IEEE/RSJ International Conference on
  Intelligent Robots and Systems}.\hskip 1em plus 0.5em minus 0.4em\relax IEEE,
  2012, pp. 2651--2652.

\bibitem{concha2016visual}
A.~Concha, G.~Loianno, V.~Kumar, and J.~Civera, ``Visual-inertial direct
  slam,'' in \emph{2016 IEEE International Conference on Robotics and
  Automation (ICRA)}.\hskip 1em plus 0.5em minus 0.4em\relax IEEE, 2016, pp.
  1331--1338.

\bibitem{ORB-SLAM2}
R.~Mur-Artal and J.~D. Tardos, ``Orb-slam2: an open-source slam system for
  monocular, stereo and rgb-d cameras,'' \emph{IEEE Transactions on Robotics},
  vol.~33, no.~5, 5, Oct 2017.

\bibitem{engel2015large}
J.~Engel, J.~St{\"u}ckler, and D.~Cremers, ``Large-scale direct slam with
  stereo cameras,'' in \emph{2015 IEEE/RSJ International Conference on
  Intelligent Robots and Systems (IROS)}.\hskip 1em plus 0.5em minus
  0.4em\relax IEEE, 2015, pp. 1935--1942.

\bibitem{von2018direct}
L.~Von~Stumberg, V.~Usenko, and D.~Cremers, ``Direct sparse visual-inertial
  odometry using dynamic marginalization,'' in \emph{2018 IEEE International
  Conference on Robotics and Automation (ICRA)}.\hskip 1em plus 0.5em minus
  0.4em\relax IEEE, 2018, pp. 2510--2517.

\bibitem{engel2014lsd}
J.~Engel, T.~Sch{\"o}ps, and D.~Cremers, ``Lsd-slam: Large-scale direct
  monocular slam,'' in \emph{European conference on computer vision}.\hskip 1em
  plus 0.5em minus 0.4em\relax Springer, 2014, pp. 834--849.

\bibitem{gao2018ldso}
X.~Gao, R.~Wang, N.~Demmel, and D.~Cremers, ``Ldso: Direct sparse odometry with
  loop closure,'' in \emph{2018 IEEE/RSJ International Conference on
  Intelligent Robots and Systems (IROS)}.\hskip 1em plus 0.5em minus
  0.4em\relax IEEE, 2018, pp. 2198--2204.

\bibitem{fischler1981random}
M.~A. Fischler and R.~C. Bolles, ``Random sample consensus: a paradigm for
  model fitting with applications to image analysis and automated
  cartography,'' \emph{Communications of the ACM}, vol.~24, no.~6, pp.
  381--395, 1981.

\bibitem{ORB}
E.~Rublee, V.~Rabaud, K.~Konolige, and G.~Bradski, ``Orb: An efficient
  alternative to sift or surf,'' \emph{IEEE International Conference on
  Computer Vision}, vol.~58, no.~11, pp. 2564--2571, 2011.

\bibitem{lowe1999object}
D.~G. Lowe \emph{et~al.}, ``Object recognition from local scale-invariant
  features.'' in \emph{iccv}, vol.~99, no.~2, 1999, pp. 1150--1157.

\bibitem{li2006five}
H.~Li and R.~Hartley, ``Five-point motion estimation made easy,'' in \emph{18th
  International Conference on Pattern Recognition (ICPR'06)}, vol.~1.\hskip 1em
  plus 0.5em minus 0.4em\relax IEEE, 2006, pp. 630--633.

\bibitem{nister2004efficient}
D.~Nist{\'e}r, ``An efficient solution to the five-point relative pose
  problem,'' \emph{IEEE transactions on pattern analysis and machine
  intelligence}, vol.~26, no.~6, pp. 0756--777, 2004.

\bibitem{longuet1981computer}
H.~C. Longuet-Higgins, ``A computer algorithm for reconstructing a scene from
  two projections,'' \emph{Nature}, vol. 293, no. 5828, p. 133, 1981.

\bibitem{GalvezTRO12}
D.~G\'alvez-L\'opez and J.~D. Tard\'os, ``Bags of binary words for fast place
  recognition in image sequences,'' \emph{IEEE Transactions on Robotics},
  vol.~28, no.~5, pp. 1188--1197, October 2012.

\bibitem{OlsonGraph2006}
E.~Olson, J.~Leonard, and S.~Teller, ``Fast iterative optimization of pose
  graphs with poor initial estimates,'' 2006, pp. 2262--2269.

\bibitem{geometrical}
R.~I.~H. B.~Triggs, P. F.~McLauchlan and A.~W. Fitzgibbon, ``Bundle adjustment
  a modern synthesis,'' \emph{International Workshop on Vision Algorithms:
  Theory and Practice}, vol. 1883, no. 1883, pp. 298--372, 1999.

\bibitem{knoblauch2011non}
D.~Knoblauch, M.~Hess-Flores, M.~A. Duchaineau, K.~I. Joy, and F.~Kuester,
  ``Non-parametric sequential frame decimation for scene reconstruction in
  low-memory streaming environments,'' in \emph{International Symposium on
  Visual Computing}.\hskip 1em plus 0.5em minus 0.4em\relax Springer, 2011, pp.
  359--370.

\bibitem{lepetit2009epnp}
V.~Lepetit, F.~Moreno-Noguer, and P.~Fua, ``Epnp: An accurate o (n) solution to
  the pnp problem,'' \emph{International journal of computer vision}, vol.~81,
  no.~2, p. 155, 2009.

\bibitem{Horn}
B.~K.~P. Horn, ``Closed-form solution of absolute orientation using unit
  quaternions,'' \emph{J. Opt. Soc. Amer. A}, vol.~4, no.~4, pp. 629--C642,
  1987.

\bibitem{bian2017gms}
J.~Bian, W.-Y. Lin, Y.~Matsushita, S.-K. Yeung, T.-D. Nguyen, and M.-M. Cheng,
  ``Gms: grid-based motion statistics for fast, ultra-robust feature
  correspondence,'' in \emph{Proceedings of the IEEE Conference on Computer
  Vision and Pattern Recognition}, 2017, pp. 4181--4190.

\bibitem{engel2016monodataset}
J.~Engel, V.~Usenko, and D.~Cremers, ``A photometrically calibrated benchmark
  for monocular visual odometry,'' in \emph{arXiv:1607.02555}, July 2016.

\bibitem{engel2018direct}
J.~Engel, V.~Koltun, and D.~Cremers, ``Direct sparse odometry,'' \emph{IEEE
  transactions on pattern analysis and machine intelligence}, vol.~40, no.~3,
  pp. 611--625, 2018.

\end{thebibliography}



\end{document}